\documentclass[journal]{IEEEtran}
\usepackage{graphicx}
\usepackage{subfigure}  % support sub-figure

\usepackage{color}
\usepackage{multirow}
\usepackage{amsmath}
\usepackage{booktabs}
\usepackage{float}
\usepackage{bm}
\usepackage{wrapfig}
\usepackage{mathrsfs}
\usepackage{bbding}
\usepackage{amssymb}
\usepackage{caption}

\usepackage{amsfonts}       % blackboard math symbols
\usepackage{nicefrac}       % compact symbols for 1/2, etc.
\usepackage{microtype}      % microtypography
\usepackage{xcolor}

\usepackage{array} 
\usepackage{pifont}

\usepackage{bbding}

\usepackage{colortbl}
\definecolor{mygray}{gray}{.9}
\definecolor{mypink}{rgb}{.99,.91,.95}
\definecolor{mycyan}{cmyk}{.3,0,0,0}

\newcommand{\ie}{\textit{i}.\textit{e}., }
\newcommand{\eg}{\textit{e}.\textit{g}., }
\newcommand{\etc}{\textit{etc}}

\usepackage[pagebackref,breaklinks,colorlinks]{hyperref}
\usepackage{wrapfig}

\ifCLASSINFOpdf
\else
\fi
\begin{document}

\title{A Unified Transformer Framework for Group-based Segmentation: Co-Segmentation, Co-Saliency Detection and Video Salient Object Detection}

\author{Yukun~Su, Jingliang~Deng, Ruizhou~Sun, Guosheng~Lin, 
        and~Qingyao~Wu% <-this % stops a space
\thanks{Y. Su, J. Deng, R. Sun and Q. Wu are with the School of Software Engineering, South China University
	of Technology, Guangzhou 510640, China, E-mail: suyukun666@gmail.com.}% <-this % stops a space
\thanks{Y. Su and G. Lin is with the School of Computer Science and Engineering, Nanyang Technological University, Singapore.}% <-this % stops a space
%\thanks{Y. Su and J. Deng contributed equally to this paper.}
\thanks{G. Lin and Q. Wu are the corresponding authors.}}

\markboth{Journal of \LaTeX\ Class Files,~Vol.~14, No.~8, March~2022}%
{Shell \MakeLowercase{\textit{et al.}}: Bare Demo of IEEEtran.cls for IEEE Journals}

\maketitle

\begin{abstract}
Humans tend to mine objects by learning from a group of images or several frames of video since we live in a dynamic world. In the computer vision area, many researches focus on co-segmentation (CoS), co-saliency detection (CoSD) and video salient object detection (VSOD) to discover the co-occurrent objects.
However, previous approaches design different networks on these similar tasks separately, and they are difficult to apply to each other, which lowers the upper bound of the transferability of deep learning frameworks. Besides, they fail to take full advantage of the cues among inter- and intra-feature within a group of images.
In this paper, we introduce a unified framework to tackle these issues, term as \textbf{UFO} (\textbf{U}nified \textbf{F}ramework for Co-\textbf{O}bject Segmentation).
Specifically, we first introduce a transformer block, which views the image feature as a patch token and then captures their long-range dependencies through the self-attention mechanism. This can help the network to excavate the patch structured similarities among the relevant objects.
Furthermore, we propose an intra-MLP learning module to produce self-mask to enhance the network to avoid partial activation.
Extensive experiments on four CoS benchmarks (PASCAL, iCoseg, Internet and MSRC), three CoSD benchmarks (Cosal2015, CoSOD3k, and CocA) and four VSOD benchmarks (DAVIS$_{16}$, FBMS, ViSal and SegV2) show that our method outperforms other state-of-the-arts on three different tasks in both accuracy and speed by using the same network architecture , which can reach 140 FPS in real-time. Code is available at \url{https://github.com/suyukun666/UFO}

\end{abstract}

\begin{IEEEkeywords}
Co-object, Long-range Dependency, Transformer, Activation.
\end{IEEEkeywords}

% \fi
\IEEEpeerreviewmaketitle

\section{Introduction}
\IEEEPARstart{O}{bject} segmentation~\cite{chen2017deeplab,chen2018encoder,su2021context} and detection~\cite{bochkovskiy2020yolov4,law2018cornernet} are the core tasks in computer vision. 
In our real world, the continuous emergence of massive group-based data and dynamic multi-frame data make deep learning more in the direction of human vision.
As a result, more and more studies focus on co-segmentation (CoS)~\cite{zhang2020deep,zhang2021cyclesegnet}, co-saliency detection (CoSD)~\cite{zhang2021deepacg,fan2021group} and video salient object detection (VSOD)~\cite{gu2020pyramid,ji2021full}.
Among them, these tasks all share the common objects with the same attributes given a group of relevant images or within several adjacent frames in the video. They all essentially aim to discover and segment the co-occurrent object by imitating the human vision system. 

\begin{figure}
	\begin{center}
		\centering
		\includegraphics[width=3.3in]{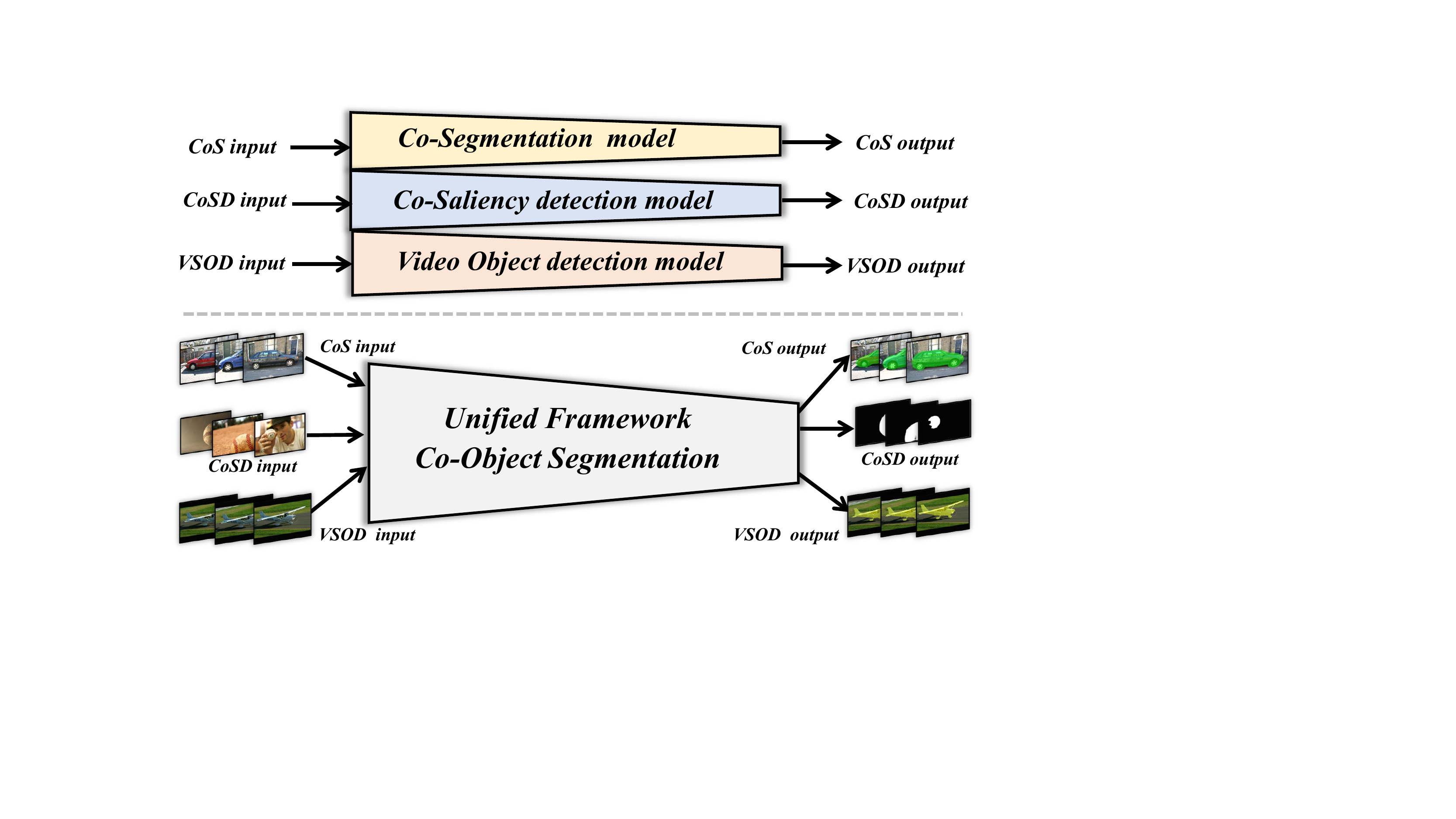}
	\end{center}
	\caption{\textbf{The main purpose of our proposed framework}. The co-object images can be fed into our network to yield the accurate masks in real-time using the same architecture.}
	\label{fig1}
	\vspace{-8pt}
\end{figure}

However, as shown in Fig~\ref{fig1} top, previous methods tend to design different networks on these tasks in isolation, which may lower the upper bound on the ease of usage of deep learning frameworks. 
Besides, the transferability and applicability of these methods are relatively poor.
For example, some of the CoS~\cite{zhang2020deep,li2019group}, CoSD~\cite{zhang2021summarize,zhang2021deepacg} and VSOD methods~\cite{gu2020pyramid,ji2021full} are well-trained on the same dataset~\cite{lin2014microsoft,wang2017learning}, they fail to achieve comparable performance on all benchmarks but only their task-targeted benchmarks. 
This illustrates that there may exist some limitations in previous approaches. 
To be specific, most of the co-object segmentation and saliency detection networks are based on matching strategies~\cite{ma2021image,liu2020semantic}, which enable the network to extract and propagate the feature representation not only to express the images individual properties but also reflect the relevance and interaction among group-based images. Some recent works~\cite{zhang2020adaptive,zhang2020deep} utilize spectral clustering~\cite{von2007tutorial} to mine the corresponding object regions. However, such methods are unstable and they are deeply dependent on the upper stream extracted features. Some  researches~\cite{zhang2021deepacg,zhang2020deepemd} adopt distance measure metric~\cite{solomon2016entropic,rubner2000earth} to model the relationship among image pixels, but they can only handle the pair-wise information and it is cumbersome to address group-wise relationships. Some others, like~\cite{fan2021group,zhang2021summarize} try to use the CNN-based attention technique to establish the relationships among group-based images. However, convolution operations produce local receptive fields and experience difficulty to capture long-range feature dependencies among pixels, which may affect the learning representation of co-object regions.
Furthermore, some of the VSOD approaches~\cite{ji2021full,xu2019video} capture the pair-wise feature between frames with the help of optical flow~\cite{ilg2017flownet}, which greatly increases the network running cost and reduces the convenience of using the networks.

To this end, we design a unified framework as shown in Fig~\ref{fig1} bottom, term as \textbf{UFO} (\textbf{U}nified \textbf{F}ramework for Co-\textbf{O}bject Segmentation), to jointly address the aforementioned drawbacks. 
%Recently, visual transformer~\cite{dosovitskiy2020image} has been proved very effective in computer vision area.
Specifically, to better reflect the relationships among group-based images, we first introduce a transformer block to insert into our network. It splits the image feature as a patch token and then captures their global dependency thanks to the self-attention mechanism and Multilayer Perceptron (MLP) structure. This can help the network learn complex spatial features and reflect long-range semantic correlations to excavate the patch structured similarities among the relevant objects.
The inherent global feature interaction capability of the visual transformer~\cite{dosovitskiy2020image} frees us from the computationally expensive similarity matrices as some previous methods~\cite{zhang2020deep}. Therefore, our method can achieve real-time performance.
In addition to improving inter-collaboration in group-based images, we also propose an intra-MLP learning module to enhance the single image.
As it is common that the encoder of the network only focuses on the most discriminative part of the objects~\cite{zhou2016learning}, in order to avoid partial activation, we add the intra-MLP operation to produce global receptive fields. For each query pixel, it will match with its top-$K$ potentially corresponding pixels, which can help the network learn divergently. Then we produce the self-masks and add them to the decoder to enhance the network.
Finally, extensive experiments on four CoS benchmarks (\ie PASCAL, iCoseg, Internet and MSRC), three CoSD benchmarks (\ie Cosal2015, CoSOD3k, and CocA) and four VSOD benchmarks (\ie DAVIS$_{16}$, FBMS, ViSal and SegV2)  demonstrate the superiority of our approach and it can outperform the state-of-the-arts in both accuracy and speed by using the same network architecture.

The main contributions of our paper are the following:
\begin{itemize}
\item We propose a unified framework for group-based image co-object segmentation (\textbf{UFO}). To the best of our knowledge, we take the early attempt to complete three different tasks (co-segmentation, co-saliency detection, video salient object detection) using the same network architecture without using additional prior. 
\item We introduce the transformer block to capture the feature long-range dependencies among group-based images through the self-attention mechanism. Besides, we design an intra-MLP learning module to avoid partial activation to further enhance the network.
\item Experiments on four CoS benchmarks, three CoSD benchmarks, and four VSOD benchmarks in different tasks show the effectiveness of our proposed method. It can achieve the new state-of-the-art performance and reach 140 FPS in real-time.
\end{itemize}

\section{Related Work}
\label{sec:rela}
\textbf{Co-Segmentation (CoS).} 
Co-Segmentation is introduced by Rother $\emph{et~al.}$~\cite{rother2006cosegmentation}, which aims to segment the common objects in pair images. Early conventional works like Gabor filters~\cite{hochbaum2009efficient} and SIFT~\cite{rubinstein2013unsupervised} try to extract low-level image features and then detect image foreground. As deep learning recently emerges and demonstrates the success in many computer vision applications, more and more recent studies 
adopt deep visual features to train object co-segmentation. Chen $\emph{et~al.}$~\cite{chen2018semantic} and Li $\emph{et~al.}$~\cite{li2018deep} first propose the siamese fully convolutional  network to solve the object co-segmentation task with a mutual correlation layer. However, both of them can not achieve satisfactory performance and they can only deal with pair-wise images.
Later, Li $\emph{et~al.}$~\cite{li2019group} and Zhang~\cite{zhang2021cyclesegnet} $\emph{et~al.}$ both propose group-wise networks using LSTM~\cite{hochreiter1997long}. Although they can improve the performance, they are computationally expensive because of the serial structure of recurrent neural network. Besides, training such methods will have a risk such as forgetting historical information. More recently, Chen $\emph{et~al.}$~\cite{chen2020show} proposes a matching strategy to jointly complete semantic matching and object Co-segmentation. Such a method is targeted for bipartite matching, which is hard to apply to group-based segmentation. Zhang~\cite{zhang2020deep} later designs a spatial-semantic network with sub-cluster optimizing. The cluster 
results are deeply dependent on the upper extracted features and thus, it may make the training unstable.

\vspace{1ex}

\textbf{Co-Saliency Detection (CoSD).} Co-Saliency Detection is similar to Co-Segmentation. The main difference between them lies in that salient detection mimics the human vision system to distinguish the most visually distinctive regions~\cite{zhang2020gradient,zhang2021deepacg}. Previous standard methods~\cite{li2013co,song2016rgbd,jerripothula2016cats} try to use hand-crafted cues or super-pixels prior to  discover the co-saliency from the images.
Later, researchers pay more attention to explore the deep-based models in a data-driven manner in various ways, \ie co-category semantic fusion~\cite{wang2019robust,zhang2020deep}, gradient-induced~\cite{zhang2020gradient} and CNN-based self-attention~\cite{fan2021group,zhang2021summarize}, \etc. However, these works do not fully consider the global correlation among the group-based images since the convolution local receptive fields. 
Some recent works~\cite{jiang2019unified,zhang2020adaptive} exploit GCN to solve the non-local problem in CNN. However, in these methods, a large number of similarity matrices and adjacency matrices need to be constructed for the graph, which will slow down the networks and are computational costly. Zhang $\emph{et~al.}$~\cite{zhang2021deepacg} proposes to use GW distance~\cite{memoli2014gromov} to build the dense correlation volumes for image pixels.
However, it has to select a target image and source images, which ignore the attention on the target image itself. Moreover, the distance metric problem in the network needs a sub-solver to optimize. This will cost more time for matching.
Compared to it, for a query patch, our transformer block can capture the relationships of both inter-pixel within group-based images and intra-pixel within a single image, which can model the global long-range semantic correlations rapidly. 

\begin{figure*}[t]
		\begin{center}
			\centering
			\includegraphics[width=6.8in]{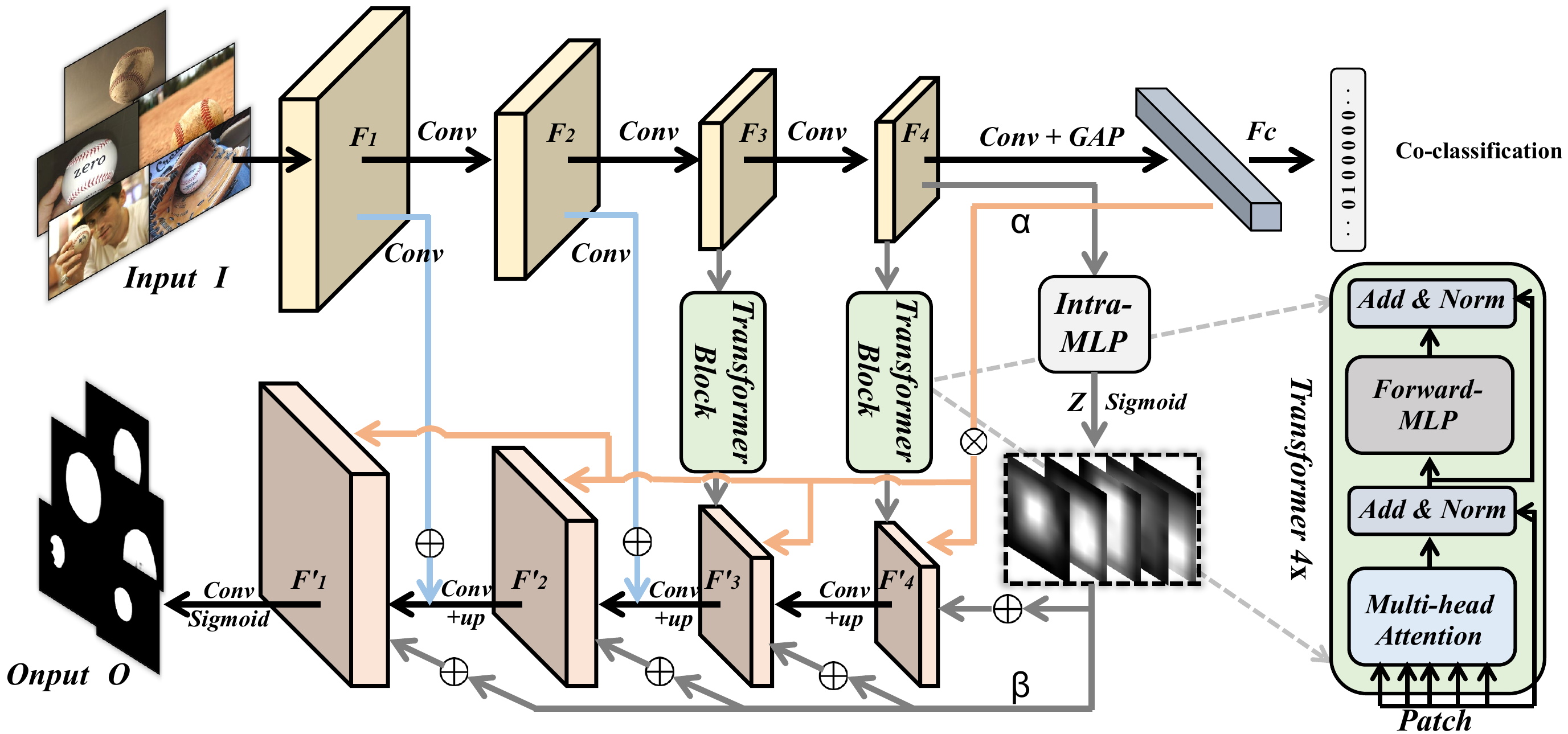}
		\end{center}
		\caption{\textbf{The pipeline of our proposed method}. The given group-based input images $\mathcal{I}$ are first fed into the encoder, yielding the multi-scale feature maps $F$. Then we employ transformer blocks (see Fig~\ref{fig3} for more details) on the last two layers to capture the images long-range correlations to model the patch structured similarities among the relevant objects, which will output the updated co-attention layers $F'$. In addition, the last layer feature are passed through a classification sub-network and an intra-MLP module. The former exploits the co-category associated information and the latter produce the self-masks. Finally, they are combined with the updated co-attention layers to enhance the co-object semantic-aware regions in the pyramid network structure-like decoder to produce the output $\mathcal{O}$.}
		\label{fig2}
\end{figure*}

\vspace{1ex}

\textbf{Video Salient Object Detection (VSOD).} In video salient object detection, since the content of each frame is highly correlated, it can be considered whose purpose is to capture long-range feature information among the adjacency frame. Traditional methods~\cite{wei2012geodesic,wang2017saliency,xu2019video1,zhou2018improving} are usually based on classic heuristics in image salient object detection area.
Subsequently, some works~\cite{tran2015learning,mahadevan2020making} rely on 3D-convolution to capture the temporal information. However, the 3D convolution operation is very time-costly.
In recent, more and more schemes~\cite{ji2021full,yan2019semi,xu2019video1} propose to combine optical flow~\cite{ilg2017flownet} to locate the representative salient objects in video frames. In practical usage, obtaining additional prior information such as optical flow will make the deep learning network inconvenient, which can not be a real sense of end-to-end network.
More recently, some approaches exploit attention-based mechanisms to better establish the pair-wise relation in the area in consecutive frames. Fan $\emph{et~al.}$~\cite{fan2019shifting} and Gu $\emph{et~al.}$~\cite{gu2020pyramid} propose a visual-attention-consistent module and a pyramid constrained self-attention block to better capture the temporal dynamics cues, respectively. However, these models can not transfer well to the above  tasks.
Therefore, in this paper, we propose a unified framework to solve these problems in a more comprehensive way.

\section{Methodology}

\subsection{Architecture Overview}
Given a group of $N$ images $\mathcal{I} = \left\{ I^n \right\}^N_{n=1}$ that contain co-occurrent objects, our target is to produce the accurate masks output $\mathcal{O} = \left\{ O^n \right\}^N_{n=1}$ that represent the share foregrounds. To achieve this goal, we propose a unified framework for co-object segmentation (\textbf{UFO}). As depicted in Fig~\ref{fig2}, the overall architecture of our network is based on an encoder (\ie VGG16~\cite{simonyan2014very}) to extract multi-scale layer features $\left\{ F_1, F2, F3, F_4 \right\}$. The transformer blocks in the last two layers are responsible for matching the co-objects similarities in multi-resolution feature maps, producing the enhanced co-attention maps $\left\{ F'_3, F'_4 \right\}$. Besides, the intra-MLP learning representation and the co-category semantic guidance are leveraged to combine with the updated maps in the decoder to enhance co-object regions. Specifically, the co-category embedding response $\alpha$ is multiplied by the decoder features, and the intra self-masks $\beta$ are added to the decoder features.
The encoder-decoder structure in our network is similar to the feature pyramid network~\cite{lin2017feature}, whose top-level features are fused by upsampling with low-level features in skip connection. 
Furthermore, the input of our network can be not only a group of images with relevant objects but also a multi-frame video.

\begin{figure*}
		\begin{center}
			\centering
			\includegraphics[width=6.8in]{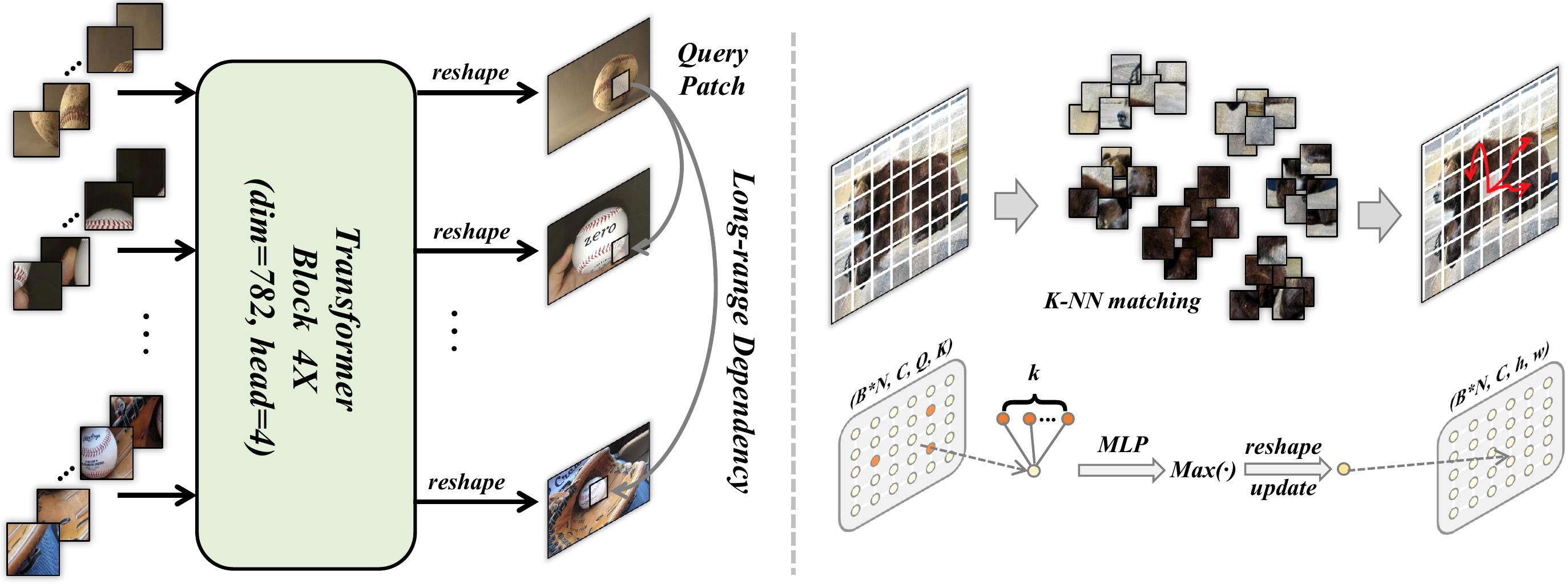}
		\end{center}
		%\vspace{-3ex}
		\caption{\textbf{Detailed illustration of the transformer block and the intra-MLP learning module}. Left: A group of images $\mathcal{I}$ are reshaped into patches. The transformer encoder will output the enhanced patch embeddings implemented by the multi-head MLP. We then reshape all the patches into image-like feature maps. Right: For each query image feature patch, it will match with its top-$K$ potentially corresponding patches. Then, it will be updated by aggregating different sub-region representations using MLP operation.}
		\label{fig3}
		%\vspace{-3ex}
	\end{figure*}

\subsection{Transformer Block}
\label{TS}

\noindent \textbf{Preliminaries.}
In visual transformer~\cite{dosovitskiy2020image}, in order to handle 2D images, an input image will be divided into a sequence of flattened 2D patches. The transformer uses a constant latent vector size $D$ in all its layers for each token embedding. Each layer consists of a multi-head self-attention layer and a Multilayer Perceptron (MLP) operation. Compared to other group-based image matching strategies and feature extraction methods, we consider that transformer block has two main advantages: (1) 
Although transformer is not specially designed for image matching, it has the intrinsic ability to capture global cues across all input patch tokens, which we have mentioned in Sec.\ref{sec:rela}. (2) The transformer parallelizes the patch tokens, which is faster than some of the serial processing methods and can achieve real-time performance.

\vspace{1ex}

\noindent \textbf{Semantic Patch Collaboration.}
\label{spc}
The detailed workflow of the transformer block is shown in Fig~\ref{fig3} left. Since we employ transformer blocks both on $F_3$ and $F_4$ layers in multi-resolution feature maps, for simplicity, we omit $F_3$ in the following. Concretely, we first reshape $F_4 \in \mathbb{R}^{B*N \times C \times h \times w}$ ($w$ and $h$ denote the spatial size of the feature map, $C$ is the feature dimension, $B$ and $N$ denote the batch size and group size, respectively) into a sequence flattened patch tokens $T \in \mathbb{R}^{B \times C \times P}$, where $P = N*h*c$. All these tokens are then fed into the transformer blocks and yield the updated patch embeddings $T'$ by: 

\begin{equation}
%\footnotesize
T' = \text{Multi-Head}(\text{MLP}(T)),\tag{1}
\label{eq1}
\end{equation}
where the trainable linear projection maps the patches from the original $C$ dimensions to $D$ dimensions, and back to $C$ dimensions.
%Note that we do not need to add the additional class token as in~\cite{dosovitskiy2020image}. 
Afterward, we reshape $T'$ back into an image-like feature map $F'_4 \in \mathbb{R}^{B*N \times C \times h \times w}$.
Due to the inductive bias~\cite{goyal2020inductive} of convolution, the features extracted by the encoder convolution layers are not sensitive to the global location of features but only care about the existence of decisive features. The proposed transformer block can complement convolution and enhance the representations of the global co-object regions.

\vspace{1ex}

\noindent \textbf{What it learns?}
Fig~\ref{fig4} shows what the transformer blocks learn (the output is from $F'_3$). As can be seen in the 3$^{rd}$ row, the corresponding image patches of the co-object regions are highly activated after operated by the transformer block comparing to the 2$^{nd}$ row, which illustrates that the attention mechanism in transformer projection function can adaptively pay attention to the targeted areas and assign more weights to capture the global cues from all the images.
In addition, we also visualize the query attention of the picked image  patch (\ie the endpoints of the “banana” and the “head” of the penguin) to show what it will match.  
The qualitative visualizations in the 3$^{rd}$ row further validate that the transformer block can help the network model the long-range dependencies among different location pixels.

\subsection{Intra-MLP module}
In addition to the image inter-collaboration, we also propose an intra-MLP learning module to activate more self-object areas within a single image. 
%alleviate this issue, 
As shown in Fig~\ref{fig3} right, the top layer feature map from the encoder are viewed as different patches. We consider that different patches will not just match with their nearest ones as in CNN (local receptive fields) since the long-distance patch features may share some similar responses (\ie color and texture).
Motivated by this, we can fuse the non-local semantic information to improve the object learning representation. Concretely, the top layer feature $F_4 \in \mathbb{R}^{B*N \times C \times h \times w}$ is reshaped into $\overline{F}_4 \in \mathbb{R}^{B*N \times C \times Q}$, where $Q = h * w$ is the number of patches. We then construct a matrix $M$ that represents the similarity of each patch within a single image. Specifically, we use ${\ell_2}$- distance to measure the relationship between the two arbitrary patches. Since we use normalized channel features, by removing the constant, the matrix $M \in \mathbb{R}^{B*N \times Q \times Q}$ can be formulated as:

\begin{equation}
%\footnotesize
M = \overline{F}_4^{\mathit{T}} \overline{F}_4.\tag{2}
\label{eq2}
\end{equation}

To avoid the patches match with themselves, the diagonal elements of the matrix are set to $-INF$. For each query patch, we perform $KNN$ operation on the matrix to select its potentially corresponding target patches. Then, it will output a tensor in $Q \times K$ shape, which indicates the patches along with their top-$K$ semantically related patches. 
After that, we can acquire the $\hat{F}_4 \in \mathbb{R}^{B*N \times C \times Q \times K}$, and then we perform MLP with $MAX(\cdot)$ operations on it to get the updated feature $Z \in \mathbb{R}^{B*N \times C \times Q}$ as:

\begin{figure*}[t]
	\begin{center}
		\centering
		\includegraphics[width=6.8in]{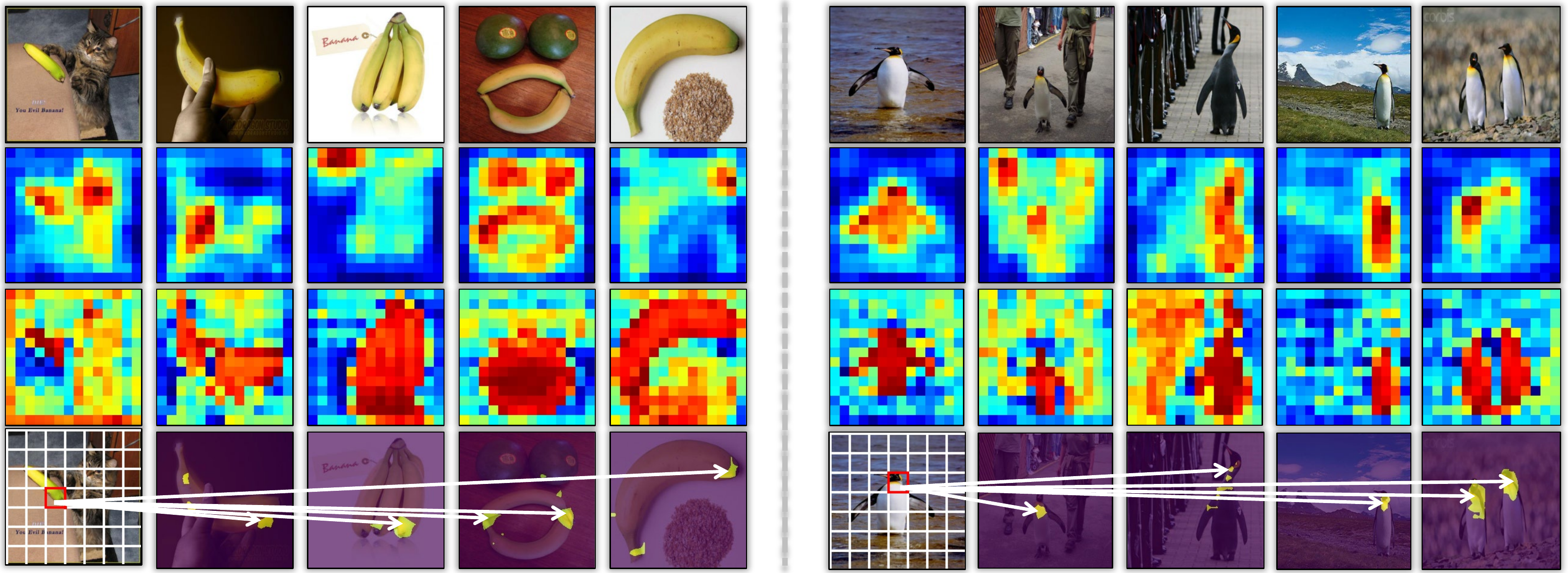}
	\end{center}
	
	\caption{\textbf{Analysis of the transformer block}. The $1^{st}$ row: original group-based images input, note that we deliberately pick some hard examples with complex backgrounds and multi-objects. The $2^{nd}$ row: the responded maps before transformer. The $3^{rd}$ row: the responded maps of the patch tokens after transformer. The $4^{th}$ row: the corresponding attention maps of the selected patches (marked with red rectangles in the query images).}
	\label{fig4}
\end{figure*}

\begin{equation}
%\footnotesize
Z = \text{MAX}(\text{MLP}(\hat{F}_4)),\tag{3}
\label{eq3}
\end{equation}
where $MAX$ is the element-wise maximum operation. This guarantees that the combination of MLP and symmetric function can arbitrarily
approximate any continuous set function~\cite{qi2017pointnet}.
The purpose of this step is to combine the target feature with its top-$K$ features appearance change information and learn through the perceptron.
Finally, we reshape $Z \in \mathbb{R}^{B*N \times C \times Q}$ back into $Z \in \mathbb{R}^{B*N \times C \times h \times w}$ and employ $Sigmoid(\cdot)$ to $Z$ to yield the self-masks $\beta$.

\subsection{Network Training}

Moreover, since the co-category information can also be used, the top layer $F_4$ is thus passed through a convolutional layer as~\cite{wang2019robust,zhang2020deep}, following a $GAP$ layer to yield a vector $\alpha$. And the $FC$ layer classifies the embedding $\alpha$ to predict the co-category labels $\hat{y}$.
Then we combine both the $\alpha$ and self-masks $\beta$ to enhance the decoder layer in a conditional normalization way~\cite{park2019semantic} modulated with learned scale and bias.
Specifically, the variables $\alpha$ are the learned scale modulation parameters that exclude the distractors in co-objects regions by using the rich semantic-aware clues. 
The variables $\beta$ serve as the bias parameters complement and highlight the object targets by endowing spatial-aware information.
The decoder leverages the skip connection structure to fuse the low-resolution layer features from the encoder.
Ultimately, it outputs the co-object masks.

\vspace{1ex}

\noindent \textbf{Objective Function.} Firstly, classification loss is used to update the gradient propagation for the semantic information as:

\begin{equation}
%\footnotesize
\mathcal{L}_{cls} = \mathcal{L}_{ce}(y,\hat{y}),\tag{4}
\label{eq4}
\end{equation}
where $\mathcal{L}_{ce}$ is the cross-entropy loss, $y$ is the ground-truth class label and $\hat{y}$ is the prediction. Besides, the Weighted Binary Cross-Entropy (WBCE) loss for pixel-wise segmentation is also adopted as:

\begin{equation}
%\footnotesize
\begin{split}
\mathcal{L}_{wbce} &= -\frac{1}{HW}\sum_{i=1}^{H}\sum_{j=1}^{W}\gamma G(i,j)log(P(i,j)) \\
& - (1-\gamma)(1-G(i,j))log(1-P(i,j)) ,
\end{split}
\tag{5}
\label{eq5}
\end{equation}
where $H$ and $W$ denote the height and width of the image. $G(i,j) \in \left\{ 0,1 \right\}$ is the ground truth mask and $P(i, j)$ is the predicted probability. $\gamma$ is the ratio of all positive pixels over all pixels in images. Moreover, similar to~\cite{qin2019basnet,fan2021group}, IoU loss is also widely used to evaluate segmentation accuracy as follows:

\begin{equation}
%\footnotesize
\mathcal{L}_{iou} =  1- \frac{\sum_{i=1}^{H}\sum_{j=1}^{W}P(i,j)G(i,j)}{\sum_{i=1}^{H}\sum_{j=1}^{W}[P(i,j)+G(i,j)-P(i,j)G(i,j)]}.\tag{6}
\label{eq6}
\end{equation}

The whole framework is optimized by integrating all the aforementioned loss functions in an end-to-end manner:

\begin{equation}
%\footnotesize
\mathcal{L}_{total} = \mathcal{L}_{cls} + \mathcal{L}_{wbce} + \mathcal{L}_{iou}.\tag{7}
\label{eq7}
\end{equation}

\section{Experiments}

\noindent \textbf{Datasets.}  Following~\cite{zhang2021cyclesegnet,zhang2020deep}, we conduct experiments on four co-segmentation benchmarks including: PASCAL-VOC~\cite{everingham2015pascal}, iCoseg~\cite{batra2010icoseg}, Internet~\cite{rubinstein2013unsupervised} and MSRC~\cite{shotton2006textonboost}. Among them, PASCAL-VOC is the most challenging dataset with 1,037 images of 20 categories. 
For co-saliency detection, our method is evaluated on the three largest and most challenging benchmark datasets, including Cosal2015~\cite{zhang2015co}, CoCA~\cite{zhang2020gradient}, and CoSOD3k~\cite{fan2021re}. The CoSOD3k is the largest evaluation benchmark for real-world co-saliency proposed recently, which has a total of 160 categories with 3,316 images. And all of them contain multiple objects against a complex background. In terms of video salient object 
detection, we benchmark our method on four public datasets, \ie DAVIS$_{16}$~\cite{perazzi2016benchmark} (30 videos training and 20 videos validation), FBMS~\cite{ochs2013segmentation} (29 training videos and 30 testing videos), ViSal~\cite{wang2015consistent} (consists of 17 video sequences for testing) and SegV2~\cite{li2013video} (consists of 13 clips for testing).

\vspace{1.0ex}

\noindent \textbf{Evaluation Metrics.} Two widely used measures, Precision ($\mathcal{P}$) and Jaccard index ($\mathcal{J}$), are used to evaluate the performance of object co-segmentation. For co-saliency detection and video salient object detection, we adopt four evaluation metrics for comparison, including the mean absolute error MAE~\cite{perazzi2012saliency}, F-measure F$_\beta$~\cite{achanta2009frequency}, E-measure E$_m$~\cite{fan2018enhanced}, and S-measure S$_m$~\cite{fan2017structure}.

\vspace{1.0ex}

\begin{table}[]
    \centering
    \scalebox{0.95}{
        \begin{tabular}{cccc|cc|cc}
			\toprule  %添加表格头部粗线
			\toprule  %添加表格头部粗线
			\multirow{2}*{\textit{Baseline}} \
		    &\multirow{2}*{$\alpha$} \ & \multirow{2}*{$\beta$} \ & \multirow{2}*{Trans} &
			\multicolumn{2}{c|}{PASCAL} &
			\multicolumn{2}{c}{iCoseg} \\
			\cline{5-8}
			& & & & $\mathcal{P}$ & $\mathcal{J}$& $\mathcal{P}$ & $\mathcal{J}$ \\
			
			\toprule  %添加表格头部粗线
			\ding{52} &   &  &  & 81.2 & 36.8 & 85.2 & 52.8 \\
			\ding{52} & \ding{52} &  &  & 92.4 & 68.1 & 95.3 & 87.9\\
			\ding{52} &  & \ding{52} & & 82.9 & 37.6 & 86.0 & 53.3\\
			\ding{52} &  &  & \ding{52} & 92.1 & 69.5 & 95.6 & 86.4\\
			\ding{52} & \ding{52} & \ding{52} &  & 92.5 & 68.6 & 95.3 & 88.1\\
			\ding{52} & \ding{52} &  & \ding{52} & 95.0 & 72.7 & 96.9& 89.8\\
			\ding{52} &  & \ding{52} & \ding{52} & 92.8 & 69.6 & 95.9& 87.1\\
			\rowcolor{mygray} \ding{52} & \ding{52} & \ding{52} & \ding{52} & \textbf{95.4} & \textbf{73.6} & \textbf{97.6}& \textbf{90.9}\\
			\bottomrule %添加表格底部粗线
	    \end{tabular}
        }
    \vspace{5pt}
        
    \caption{Analysis of different modules and their combinations.}\label{table1}
\end{table}

\begin{table}[t]
    \centering
    \scalebox{0.95}{
        \begin{tabular}{cc|cc|cc}
        		\toprule  %添加表格头部粗线
        		\toprule  %添加表格头部粗线
        		\multicolumn{2}{c|}{\multirow{2}*{\text{Transformer}}}
        	    &
        		\multicolumn{2}{c|}{PASCAL} &
			    \multicolumn{2}{c}{iCoseg}\\
        		\cline{3-6}
        		 ~& ~& $\mathcal{P}$ & $\mathcal{J}$& $\mathcal{P}$ & $\mathcal{J}$ \\
        		
        		\toprule  %添加表格头部粗线
        		\multirow{3}*{\text{Block = 2}} \ \ & \text{Head = 2} \ \ & 94.0 & 71.8 & 95.9& 89.1 \\
        		~ & \text{Head = 4} \ \ & 94.2 & 72.5 & 96.3 & 89.4\\
        		~ & \text{Head =  6} \ \ & 94.6 & 73.0 & 96.4 & 89.9\\
        		\midrule
        		\multirow{3}*{\text{Block = 4}} \ \ & \text{Head = 2} \ \ & 94.8 & 72.9 & 97.3 & 90.4\\
        		 ~ &  \text{Head = 4} \ \ & \cellcolor{mygray} 95.4 & \cellcolor{mygray} \textbf{73.6} & \cellcolor{mygray} \textbf{97.6}& \cellcolor{mygray} \textbf{90.9}\\
        		~ & \text{Head = 6} \ \ & 95.0 & 73.5 & 97.5 & 90.2\\
        		\midrule
        		\multirow{3}*{\text{Block = 6}} \ \ & \text{Head = 2} \ \ & 95.1 & 72.9 & 97.1 & 90.6\\
        		~ & \text{Head = 4} \ \ & \textbf{95.6} & 72.8 & 97.3 & 90.2\\
        		~ &  \text{Head = 6} \ \ & 95.0 & 73.1 & 97.5 & 90.4\\
        		\bottomrule %添加表格底部粗线
                \end{tabular}
        }
        \vspace{5pt}
		\caption{Analysis of the block and multi-head number in transformer.}\label{table2}
\end{table}

\noindent \textbf{Training Details.}
The input image group $\mathcal{I}$ contains $N = 5$ images. The mini-batch size is set to 8 $\times N$.
For fair comparisons, we strictly follow the same settings as~\cite{zhang2020gradient,fan2021group}  to use VGG16~\cite{simonyan2014very} as backbone and the images are all resized to 224 $\times 224$ for training and testing unless otherwise stated.
We use 4 transformer blocks with 4 multi-heads and $K$ is set to 4 by default. And the feature dimension of the transformer linear projection function is 782.
\textbf{(1):} On both co-segmentation (CoS) and co-saliency detection (CoSD) tasks, we follow~\cite{wang2019robust,zhang2021summarize} to use the COCO-SEG~\cite{lin2014microsoft} for training, which contains 200, 000 images belonging to 78 groups. And each image has a manually-labeled binary mask with co-category labels. We leverage the Adam algorithm~\cite{kingma2014adam} as the optimization strategy with $\beta _1$ = 0.9 and $\beta_2$ = 0.999. The learning rate is initially set to 1e-5 and reduces by a half every 25,000 steps. The whole training takes about 20 hours for total 100,000 steps.
\textbf{(2):} On video salient detection task (VSOD), the common practice for most methods ~\cite{gu2020pyramid,ji2021full} are first pre-trained on the static saliency dataset. Following this scheme, we first load the weights pre-trained on CoS and CoSD tasks and use DUT~\cite{wang2017learning} dataset to train our network to avoid over-fitting. And then we frozen the co-classification layer since we do not have class labels in VSOD task. Lastly, we train on the training set of DAVIS$_{16}$ (30 clips) and FBMS (29 clips) as~\cite{ji2021full} and it takes about 4 hours.
All the experiments are conducted on a RTX 3090 GPU.
For all the runtime analysis, we report the results tested on Titan Xp GPU as in~\cite{gu2020pyramid} for fair comparisons.

\begin{table}[t]
    \centering
    \scalebox{1.15}{
        \begin{tabular}{c|cc|cc}
        		\toprule  %添加表格头部粗线
        		\toprule  %添加表格头部粗线
        		\multirow{2}*{\text{Method}} 
        	    &
        		\multicolumn{2}{c|}{PASCAL}  &
			    \multicolumn{2}{c}{iCoseg} \\
        		\cline{2-5}
        		 & $\mathcal{P}$ & $\mathcal{J}$& $\mathcal{P}$ & $\mathcal{J}$ \\
        		
        		\toprule  %添加表格头部粗线
        		\text{$F_3$ only} & 92.1 & 69.8 & 94.6 & 86.4 \\
        		\text{$F_4$ only} & 94.1 & 72.8 & 96.2 & 88.7 \\
        		\rowcolor{mygray} \textbf{$F_3$ and $F_4$} & \textbf{95.4} & \textbf{73.6} & \textbf{97.6} & \textbf{90.9}\\
        		\bottomrule %添加表格底部粗线
                \end{tabular}
        }
		\vspace{6pt}
		\caption{Analysis of the features from different stages in the encoder.}\label{table3}
\end{table}

\begin{table}[t]
    \centering
    \scalebox{1.15}{
        \begin{tabular}{c|cc|cc}
        		\toprule  %添加表格头部粗线
        		\toprule  %添加表格头部粗线
        		\multirow{2}*{\text{Method}} 
        	    &
        		\multicolumn{2}{c|}{PASCAL}  &
			    \multicolumn{2}{c}{iCoseg} \\
        		\cline{2-5}
        		 & $\mathcal{P}$ & $\mathcal{J}$& $\mathcal{P}$ & $\mathcal{J}$ \\
        		
        		\toprule  %添加表格头部粗线
        		\text{Conv} & 94.8 & 72.3 & 96.8 & 89.7 \\
        		\text{Non-Local~\cite{wang2018non}} & 95.1 & 72.9 & 97.2 & 90.1 \\
        		\rowcolor{mygray} \textbf{Intra-MLP} & \textbf{95.4} & \textbf{73.6} & \textbf{97.6} & \textbf{90.9}\\
        		\bottomrule %添加表格底部粗线
                \end{tabular}
        }
		\vspace{6pt}
		\caption{Analysis of different operations for self-masks production.}\label{table4}
\end{table}

\begin{table}[!bt]
    \centering
    \scalebox{1.15}{
        \begin{tabular}{ccc}
				\toprule  %添加表格头部粗线
				\toprule  %添加表格头部粗线
				Method  & Time (\textit
				{ms}) \\
				\midrule  %添加表格中横线
				Cluster~\cite{zhang2020deep} & 49.1  \\
				Graph~\cite{zhang2020adaptive} & 50.2  \\
				GW-Distance~\cite{zhang2021deepacg} & \textgreater 50 \\
				\rowcolor{mygray} \textbf{Transformer}  & \textbf{5.4}\\
				\bottomrule %添加表格底部粗线
		    \end{tabular}
        }
		\vspace{5pt}
		\caption{Analysis of different methods' matching time cost.}\label{table5}
    \vspace{-10pt}
\end{table}

\begin{figure*}[!hbt]
    \begin{center}
    	\centering
    	\includegraphics[width=6.8in]{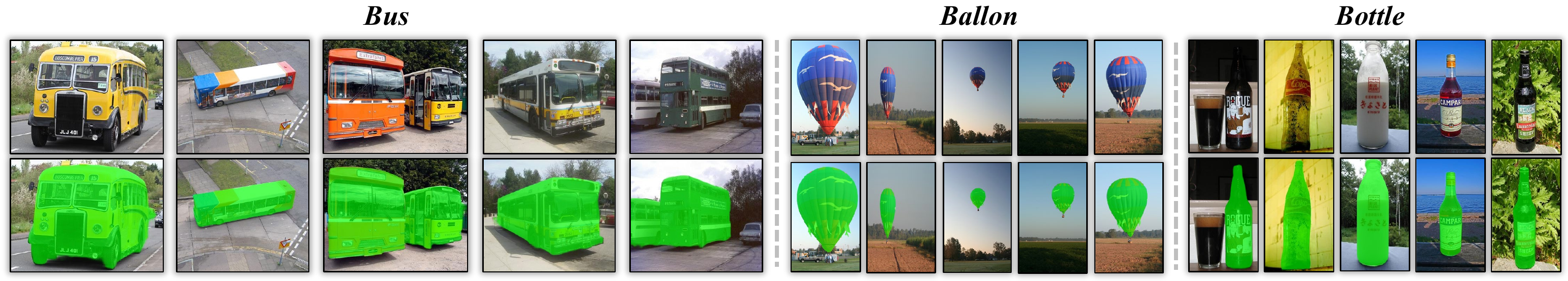}
    \end{center}
    %\vspace{-3ex}
    \caption{Qualitative results of our proposed network for “Bus”, “Ballon” and “Bottle” objects.}
    \label{fig5}
\end{figure*}

\begin{table*}[t]
    \renewcommand\arraystretch{1.3}
	\begin{center}
		\scalebox{1.0}{
			\begin{tabular}{ccc|cc|cc|cc|cc}
			
				\toprule  %添加表格头部粗线
				\multirow{2}*{Method} & 
				\multirow{2}*{Backbone} &
				\multirow{2}*{Size} &
				\multicolumn{2}{c|}{PASCAL} &
				\multicolumn{2}{c|}{iCoseg} &
				\multicolumn{2}{c|}{Internet} &
				\multicolumn{2}{c}{MSRC} \\
				\cline{4-11}
				{~}&{~}&{~}&
				$\mathcal{P}$ & $\mathcal{J}$ & 
				$\mathcal{P}$ & $\mathcal{J}$ & 
				$\mathcal{P}$ & $\mathcal{J}$ & 
				$\mathcal{P}$ & $\mathcal{J}$  \\
				% \cmidrule{2-7} & ZSL \ &  GZSL \ &  ZSL \ & GZSL \ &  ZSL \ &  GZSL \ \\
				\toprule  %添加表格头部粗线
				Jerripothula $\emph{et~al.}$\cite{jerripothula2016image}$_{\text{TMM'2016}}$& - \ & - \ &  85.2 & 45.0 & 91.9  & 72.0 & 88.9 & 64.0 & 88.7 & 71.0 \\
				
				Jerripothula $\emph{et~al.}$\cite{jerripothula2017object}$_{\text{CVPR'2017}}$& - \ & - \ &  80.1 & 40.0 & -  & - & - & - & - & - \\
				
				Wang $\emph{et~al.}$\cite{wang2017multiple}$_{\text{TIP'2017}}$& ResNet50 \ & 300 $\times$ 300 \ &  84.3 & 52.0 & 93.8  & 77.0 & - & - & 90.9 & 73.0 \\
				
				Hsu $\emph{et~al.}$\cite{hsu2018co}$_{\text{IJCAI'2018}}$& VGG16 \ & 384 $\times$ 384 \ &  91.0 & 60.0 & 96.5  & 84.0 & 92.3 & 69.8 & - & - \\
				
				Chen $\emph{et~al.}$\cite{chen2018semantic}$_{\text{ACCV'2018}}$& VGG16 \ & 512 $\times$ 512 \ &  - & 59.8 & -  & 84.0 & - & 73.1 & - & 77.7 \\
				
				Li $\emph{et~al.}$\cite{li2018deep}$_{\text{ACCV'2018}}$& VGG16 \ & 512 $\times$ 512 \ &  94.2 & 64.5 & 95.1  & 84.2 & 93.5 & 72.6 & 95.2 & 82.9 \\
				
				CARNN\cite{li2019group}$_{\text{ICCV'2019}}$& VGG19 \ & 224 $\times$ 224 \ &  94.1 & 63.0 & {\color{blue}97.9}  & 89.0 & {\color{red}97.1} & {\color{blue}84.0} & - & - \\
				
				SSNM\cite{zhang2020deep}$_{\text{AAAI'2020}}$& VGG16 \ & 224 $\times$ 224 \ &  93.7 & 66.0 & 96.5 & 88.0 & 92.3 & 67.0 & 94.3 & 76.3 \\
				
				Chen $\emph{et~al.}$\cite{chen2020show}$_{\text{TPAMI'2020}}$& ResNet101 \ & 240 $\times$ 240 \ &  93.9 & 61.0 & -  & - & 93.5 & 70.0 & - & - \\
				
				CycleSegNet\cite{zhang2021cyclesegnet}$_{\text{TIP'2021}}$& ResNet34 \ & 512 $\times$ 512 \ &  {\color{blue}96.8} & 73.6 & - & {\color{blue}92.1} & - & {\color{red}86.2} & {\color{blue}97.6} & {\color{red}89.6} \\
				
				GCoNet$\dag$\cite{fan2021group}$_{\text{CVPR'2021}}$& VGG16 \ & 224 $\times$ 224 \ &  93.5 & 69.2 & 96.9  & 89.7 & 92.5 & 70.5 & 94.0 & 80.8 \\
				
				CADC$\dag$\cite{zhang2021summarize}$_{\text{ICCV'2021}}$& VGG16 \ & 256 $\times$ 256 \ &  94.1 & 71.8 & 97.1  & 90.2 & 92.8 & 71.9 & 94.4 & 81.6 \\

				\midrule  %添加表格中横线
				%  \cellcolor{white}
	             \textbf{UFO \textit{(Ours)}} & VGG16 \ & 224 $\times$ 224 \ & \textbf{95.4} & \textbf{{\color{blue}73.6}} & \textbf{97.6} & \textbf{90.9} & \textbf{93.3} & \textbf{73.7} & \textbf{95.8} & \textbf{83.2} \\
				\textbf{UFO \textit{(Ours)}} & VGG16 \ & 256 $\times$ 256 \ & \textbf{{\color{red}96.9}} & \textbf{{\color{red}75.7}} & \textbf{{\color{red}98.1}} & \textbf{{\color{red}92.3}} & \textbf{{\color{blue}95.2}} & \textbf{74.6} & \textbf{{\color{red}97.8}} & \textbf{{\color{blue}84.3}} \\
				\bottomrule %添加表格底部粗线
		\end{tabular}}    
	\end{center}\caption{Comparisons of our method with the other state-of-the-arts on CoS datasets. $\dag$ denotes the results using the publicly released code to re-complete. The best two results on each dataset are shown in {\color{red} red} and {\color{blue} blue}.}\label{table6}
	%\vspace{-6ex}
\end{table*}

\subsection{Ablation Studies} 
To explore each component of our proposed method, we conduct several ablation studies on the CoS (PASCAL-VOC and iCoseg) datasets to demonstrate their effectiveness. 

\vspace{1ex}

\noindent \textbf{Comparisons to Baseline:}
In Tab~\ref{table1}, we investigate the effect of different proposed modules including co-category information $\alpha$, intra-MLP masks $\beta$, transformer blocks and their combinations. As can be seen, each module can boost the performance of the baseline model to different degrees.
Specifically, the proposed transformer block provides the main contributions to help improve the network performances. Whether used alone or in combination with the other two modules, it can outperform other alternatives by a large margin. This illustrates the effectiveness of each component in our framework. 

\vspace{1ex}

\noindent \textbf{Settings in Transformer:}
Tab~\ref{table2} shows the exploration of using different block and multi-head numbers in the transformer. We find that network performance can be improved when increasing the number of blocks and the multi-head attention mechanisms. However, keep increasing both the block and head numbers does not always bring gains to the network (\ie block = 6), which will also cost much computational resources.
We conjecture that large block and multi-head numbers will bring more parameter and redundant information for network learning. Therefore, we set both block and multi-head numbers to 4 in our paper.

\vspace{1ex}

\noindent \textbf{Multi-Scale Features:}
In our method, we adopt transformer blocks on different top-level layers (\ie $F_3$ and $F_4$). Since the low-level (\ie $F_1$ and $F_2$) feature maps are large, we do not consider to using them. Tab~\ref{table3} reveals that using both the features from the last two layers is better than using the features alone. This indicates that different level features can provide some vital \textbf{coarse-to-fine} information in the co-object segmentation task.

\vspace{1ex}

\noindent \textbf{Self-Mask Production:}
In Tab~\ref{table4}, we compare our intra-MLP learning module to standard convolution operation and non-local~\cite{wang2018non}. We can observe that both our method and non-local can improve the performance by accepting more receptive fields to activate object regions compared to the standard convolution. Moreover, although non-local operation can improve the original convolution, it essentially uses some $1 \times 1$ convolutions to extract features and then reshape and multiply to get the output, which is still worse than our method.

\vspace{1ex}

\noindent \textbf{Matching Time:}
We also exhibit the matching time of different methods in co-object regions mining in Tab~\ref{table5}. The results illustrate that our proposed method can not only better model co-object long-distance similarities, but also achieve the best performance in speed, which further validates the claim of the advantage of transformer block. More other analysis can be referred to the supplementary material.

\begin{figure*}[h]
		\begin{center}
			\centering
			\includegraphics[width=6.8in]{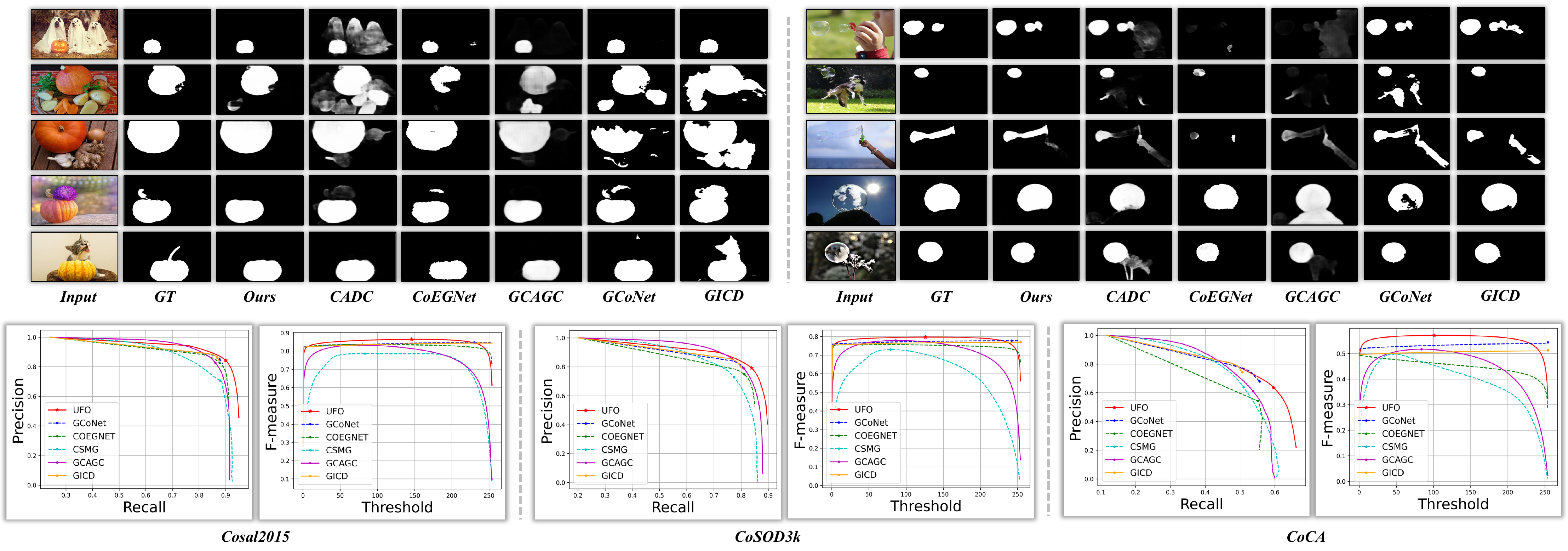}
		\end{center}
		\caption{Qualitative results of our method compared with other state-of-the-art methods and the PR and F-measure curves on three benchmark datasets.}
		\label{fig6}
\end{figure*}
	
\begin{table*}[htb!]
    \renewcommand\arraystretch{1.3}
	\begin{center}
		\scalebox{1.0}{
			\begin{tabular}{c|c|cccc|cccc|cccc}
				\toprule  %添加表格头部粗线
				\multirow{2}*{Method} & \multirow{2}*{Type} & \multicolumn{4}{c|}{Cosal2015} & \multicolumn{4}{c|}{CoSOD3k} &  \multicolumn{4}{c}{CoCA} \\
				\cline{3-14}
				{~}&{~}&
				\textit{MAE$\downarrow$} & \textit{$S_m\uparrow$} & \textit{$E_m\uparrow$} & \textit{$F_{\beta}\uparrow$} &
				\textit{MAE$\downarrow$} & \textit{$S_m\uparrow$} & \textit{$E_m\uparrow$} & \textit{$F_{\beta}\uparrow$} &
				\textit{MAE$\downarrow$} & \textit{$S_m\uparrow$} & \textit{$E_m\uparrow$} & \textit{$F_{\beta}\uparrow$}  \\
				% \cmidrule{2-7} & ZSL \ &  GZSL \ &  ZSL \ & GZSL \ &  ZSL \ &  GZSL \ \\
				\toprule  %添加表格头部粗线
				BASNet\cite{qin2019basnet}$_{\text{CVPR'2019}}$& Sin  &  0.097 & 0.820 & 0.846 & 0.784  & 0.122 & 0.753 & 0.791 & 0.696 & 0.195 & 0.589 & 0.623 & 0.397 \\
				
				PoolNet\cite{liu2019simple}$_{\text{CVPR'2019}}$& Sin  &  0.094 & 0.820 & 0.851 & 0.785  & 0.120 & 0.763 & 0.797 & 0.704 & 0.179 & 0.599 & 0.631 & 0.401 \\
				
				EGNet\cite{zhao2019egnet}$_{\text{ICCV'2019}}$& Sin  &  0.099 & 0.818 & 0.842 & 0.782  & 0.119 & 0.762 & 0.796 & 0.703 & 0.179 & 0.594 & 0.637 & 0.389 \\
				
				SCRN\cite{wu2019stacked}$_{\text{ICCV'2019}}$& Sin  &  0.097 & 0.814 & 0.854 & 0.789  & 0.118 & 0.773 & 0.806 & 0.717 & 0.166 & 0.610 & 0.658 & 0.416 \\
			
				\midrule  %添加表格中横线
				
				RCAN\cite{li2019detecting}$_{\text{IJCAI'2019}}$& Co  &  0.126 & 0.779 & 0.842 & 0.764 & 0.130 & 0.744 & 0.808 & 0.688 & 0.160 & 0.616 & 0.702 & 0.422 \\
				
				CSMG\cite{zhang2019co}$_{\text{CVPR'2019}}$& Co  &  0.130 & 0.774 & 0.818 & 0.777  & 0.157 & 0.711 & 0.723 & 0.645 & 0.124 & 0.632 & 0.734 & 0.503 \\
				
				GICD\cite{zhang2020gradient}$_{\text{ECCV'2020}}$& Co  &  0.071 & 0.842 & 0.884 & 0.834  & 0.089 & 0.778 & 0.831 & 0.743 & 0.125 & 0.658 & 0.701 & 0.513 \\
				
				SSNM\cite{zhang2020deep}$_{\text{AAAI'2020}}$& Co  &  0.102 & 0.788 & 0.843 & 0.794  & 0.120 & 0.726 & 0.756 & 0.675 & 0.116 & 0.628 & 0.741 & 0.482 \\
				
				GCAGC\cite{zhang2020adaptive}$_{\text{CVPR'2020}}$& Co  &  0.085 & 0.817 & 0.866 & 0.813  & 0.100 & 0.785 & 0.816 & 0.740 & 0.111 & 0.669 & 0.754 & 0.523 \\

				CoEGNet\cite{fan2021re}$_{\text{TPAMI'2021}}$& Co  & 0.077 & 0.836 & 0.882 & 0.832  & 0.092 & 0.762 & 0.825 & 0.736 &  0.106 & 0.612 & 0.717 & 0.493\\
				
				DeepACG\cite{zhang2021deepacg}$_{\text{CVPR'2021}}$& Co  &  {\color{red}0.064} & 0.854 & {\color{blue}0.892} & 0.842  & 0.089 & 0.792 & 0.838 & 0.756 & {\color{blue}0.102} & {\color{blue}0.688} & {\color{blue}0.771} & {\color{blue}0.552} \\
				
				GCoNet\cite{fan2021group}$_{\text{CVPR'2021}}$& Co  &  {\color{blue}0.068} & 0.845 & 0.887 & 0.847  & {\color{red}0.071} & {\color{blue}0.802} & {\color{blue}0.860} & {\color{blue}0.777} & 0.105 & 0.673 & 0.760 & 0.544 \\
				
				CADC$\dag$\cite{zhang2021summarize}$_{\text{ICCV'2021}}$& Co  &  {\color{red}0.064} & {\color{red}0.866} & {\color{red}0.906} & {\color{blue}0.862}  & 0.096 & 0.801 & 0.840 & 0.759 & 0.132 & 0.681 & 0.744 & 0.548 \\
				
				%CycleSegNet$\dag$\cite{zhang2021cyclesegnet}$_{\text{TIP'2021}}$& Co  &  - & - & - & -  & - & - & - & - & - & - & - & - \\
					
				\midrule  %添加表格中横线
				%  \cellcolor{white}
				\textbf{UFO \textit{(Ours)}} & Co & \textbf{{\color{red}0.064}} & \textbf{{\color{blue}0.860}} & \textbf{{\color{red}0.906}} & \textbf{{\color{red}0.865}}  & \textbf{{\color{blue}0.073}} & \textbf{{\color{red}0.819}} & \textbf{{\color{red}0.874}} & \textbf{{\color{red}0.797}}  & \textbf{{\color{red}0.095}} & \textbf{{\color{red}0.697}} & \textbf{{\color{red}0.782}} & \textbf{{\color{red}0.571}} \\
				\bottomrule %添加表格底部粗线
		\end{tabular}}    
	\end{center}\caption{Comparisons of our method with the other state-of-the-arts on CoSD datasets. “Sin” and “Co” denote single and co-object image saliency object detection methods, respectively. $\dag$ denotes the results using a larger input size (256 $\times$ 256) and copy and paste augmentation strategy. The best two results on each dataset are shown in {\color{red} red} and {\color{blue} blue}.}\label{table7}
	%\vspace{-6ex}
\end{table*}

\subsection{Comparison with State-of-the-arts}

\noindent \textbf{Co-Segmentation.} Tab~\ref{table6} presents the comparisons of our method with other existing CoS methods. Note that there is no standard specification for unifying the various methods on this task, and thus, we point out the respective backbone and input resolution in the table. By using the VGG16~\cite{simonyan2014very} backbone, our framework outperforms all the other methods using the same backbone, even though some of them use a larger input size. Besides, compared to the stronger backbone methods like Chen $\emph{et~al.}$~\cite{chen2020show} and CycleSegNet~\cite{zhang2021cyclesegnet}, we can also achieve relatively comparable performance and even outperforms them. CARNN~\cite{li2019group} adopts the VGG19 backbone and achieves the top 
precision on the Internet dataset. However, it is trained on the full COCO dataset. It contains 9k images belonging to 118 groups, which is much larger than our training set.
Furthermore, we also report the performance of some state-of-the-art methods in the co-saliency detection task (\ie GCoNet~\cite{fan2021group} and CADC~\cite{zhang2021summarize}). The results show that our method can also outperform them, which reflects that the transferability of such co-saliency detection methods is poor and they are unsuitable for co-segmentation.
In general, our method achieves 5 best results and 2 second-best results on 4 benchmarks, which is competitive. Fig~\ref{fig5} shows some qualitative results and more visualization results can be referred to the supplementary.

\begin{figure*}[t]
\begin{center}
	\centering
	\includegraphics[width=6.8in]{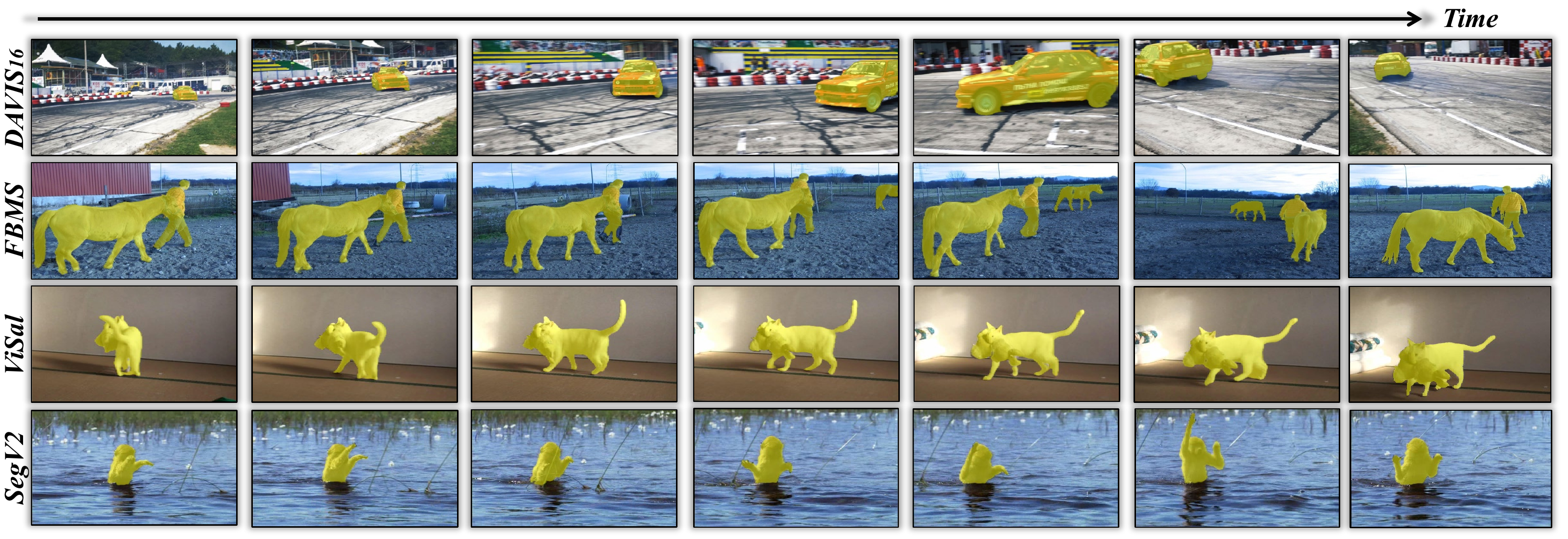}
\end{center}
%\vspace{-3ex}
\caption{Qualitative results of our proposed framework on DAVIS, FBMS, ViSal and SegV2 datasets, respectively. More visualizations can be referred to the supplementary material.}
\label{fig7}
%\vspace{-5pt}
\end{figure*}

\begin{table*}[t]
    \renewcommand\arraystretch{1.3}
	\begin{center}
		\scalebox{0.80}{
			\begin{tabular}{c|c|c|c|c|ccc|ccc|ccc|ccc}
				\toprule  %添加表格头部粗线
				\multirow{2}*{Method} & \multirow{2}*{Setting} & \multirow{2}*{\textbf{\textit{OF}}} &
				\multirow{2}*{\textbf{\textit{Sup}}} &
				\multirow{2}*{\textbf{\textit{RT}}} &
				\multicolumn{3}{c|}{DAVIS\text{$_{16}$}} & \multicolumn{3}{c|}{FBMS} &  \multicolumn{3}{c|}{ViSal} &
				\multicolumn{3}{c}{SegV2}
				\\
				\cline{6-17}
				{~}&{~}&{~}&{~}&{~}&
				\textit{MAE$\downarrow$} & \textit{$S_m\uparrow$} & \textit{$F_{\beta}\uparrow$} &
				\textit{MAE$\downarrow$} & \textit{$S_m\uparrow$} & \textit{$F_{\beta}\uparrow$} &
				\textit{MAE$\downarrow$} & \textit{$S_m\uparrow$} & \textit{$F_{\beta}\uparrow$} &
				\textit{MAE$\downarrow$} & \textit{$S_m\uparrow$} & \textit{$F_{\beta}\uparrow$}  \\
				% \cmidrule{2-7} & ZSL \ &  GZSL \ &  ZSL \ & GZSL \ &  ZSL \ &  GZSL \ \\
				\toprule  %添加表格头部粗线
				
				SCOM\cite{chen2018scom}\tiny{$_{\text{TIP'2018}}$} & DCL\cite{li2016deep} & \ding{52} & U & 38.8 &  0.048 & 0.832  & 0.783  & 0.079 & 0.794 & 0.797 & 0.122 & 0.762 & 0.831 & 0.030 & 0.815 & 0.764 \\
				
				MBNM\cite{li2018unsupervised}\tiny{$_{\text{ECCV'2018}}$}& DeepLab & \ding{52} & U & 2.6 &  0.031 & 0.887  & 0.861  & 0.047 & 0.857 & 0.816 & 0.020 & 0.898 & 0.883 & 0.026 & 0.809 & 0.716 \\
				
				PDBM\cite{song2018pyramid}\tiny{$_{\text{ECCV'2018}}$}& Res50-473 & \XSolidBrush& U &0.05 &  0.028 & 0.882 & 0.855  & 0.064 & 0.851  & 0.821 & 0.032 & 0.907  & 0.888 & 0.024 & 0.864 & 0.800 \\

				SRP\cite{cong2019video}\tiny{$_{\text{TIP'2019}}$}& - & \ding{52} & U & 17.0 &  0.070 & 0.662  & 0.660  & 0.134 & 0.648  & 0.671 & 0.092 & - & 0.752 & 0.095 & - & 0.683\\
				
				MESO\cite{xu2019video}\tiny{$_{\text{TMM'2019}}$}& - & \ding{52} & U & 50.3 & 0.070 & 0.718  & 0.660  & 0.134 & 0.635  & 0.618 & - & -  & - & - & - & -\\
				
				LTSI\cite{chen2019improved}\tiny{$_{\text{TIP'2019}}$}& VGG16-500 & \ding{52} & U & 1.4 & 0.034 & 0.876  & 0.850  & 0.087 & 0.805  & 0.799 & 0.027 & 0.922 & 0.909 & 0.028 & 0.827 & 0.862\\
				
				RSE\cite{xu2019video1}\tiny{$_{\text{TCSVT'2019}}$}& - & \ding{52} & U & 48.2 & 0.063 & 0.748  & 0.698  & 0.128 & 0.670  & 0.652 & - & - & - & - & - & -\\
				
				SSAV\cite{fan2019shifting}\tiny{$_{\text{CVPR'2019}}$}& Res50-473 & \XSolidBrush& U & 0.05 & 0.028 & 0.893  & 0.861  & {0.040} & {0.879} & 0.865 & 0.020 & 0.943 & 0.939 & 0.023 & 0.851 & 0.801\\
				
				RCR\cite{yan2019semi}\tiny{$_{\text{ICCV'2019}}$}& Res50-448 & \ding{52} & S & 0.04 & 0.027 & 0.886 & 0.848  & 0.053 & 0.872  & 0.859 & - & -  & - & - & - & -\\
				
				%SPD~\cite{}& - & - &  0.061 & 0.783 & 0.763  & 0.125 & 0.691 & 0.686 & - & -  & - & 0.049 & - & 0.890\\
				
				%3DC\cite{mahadevan2020making}\tiny{$_{\text{BMVC'2020}}$}& Res152-480 & & 0.8 &  \textbf{{\color{red}0.015}} & - & \textbf{{\color{red}0.918}}  & 0.048 & - & 0.845 & 0.019 & -  & 0.922 & - & - & -\\
				
				CAS\cite{ji2020casnet}\tiny{$_{\text{TNNLS'2020}}$}& Res50 & \XSolidBrush& U & - &  0.032 & 0.873 & 0.860  & 0.056 & 0.856 & 0.863 & - & -  & - & 0.029 & 0.820 & 0.847\\

				PCSA\cite{gu2020pyramid}\tiny{$_{\text{AAAI'2020}}$}& MobV3-448 & \XSolidBrush& U &{\color{blue}0.009} &  0.022 & 0.902 &  0.880  & 0.040 & 0.868  & 0.837 & 0.017 & 0.946 & {\color{blue}0.940} & 0.025 & 0.865 & 0.810\\
				
				DFNet\cite{zhen2020learning}\tiny{$_{\text{ECCV'2020}}$}& DeepLab & \XSolidBrush& U & - &  {\color{blue}0.018} & - & 0.899  & 0.054 & - & 0.833 & 0.017 & - & 0.927 & - & - & -\\
				
				FSNet\cite{ji2021full}\tiny{$_{\text{ICCV'2021}}$}& Res101-352 & \ding{52} & U & 0.08 & 0.020 & {\color{red}0.920}  & {\color{red}0.907}  & 0.041 & 0.890 & {\color{blue}0.888} & - & - & - & 0.023 & 0.870 & 0.772\\
				
				%RTNet\cite{ren2021reciprocal}\tiny{$_{\text{CVPR'2021}}$}& Res34-352 & \ding{52} & - & - & -  & -  & - & - & - & - & - & - & - & - & -\\
				
				ReuseVOS$\dag$\cite{park2021learning}\tiny{$_{\text{CVPR'2021}}$}& Res18-480 & \XSolidBrush & S & 0.02 & 0.019 & 0.883  & 0.865  & 0.027 & 0.888 & 0.884 & 0.020 & 0.928 & 0.933 & 0.025 & 0.844 & 0.832\\
				
				TransVOS$\dag$\cite{mei2021transvos}\tiny{$_{\text{PrePrint'2021}}$}& Res50-240 & \XSolidBrush & S & 0.06 & {\color{blue}0.018} & 0.885  & 0.869  & 0.038 & 0.867 & 0.886 & 0.021 & 0.917 & 0.928 & 0.024 & 0.816 & 0.800\\
					
				\midrule  %添加表格中横线
				%  \cellcolor{white}
				\textbf{UFO \textit{(Ours)}} & VGG16-224 & \XSolidBrush& U & \textbf{{\color{red}0.007}} & 0.036 & 0.864  & 0.828 & \textbf{{\color{red}0.028}} & \textbf{{\color{red}0.894}} & \textbf{{\color{red}0.890}}  & \textbf{{\color{red}0.011}} & \textbf{{\color{blue}0.953}}  & \textbf{{\color{blue}0.940}} & \textbf{{\color{blue}0.022}} & \textbf{{\color{blue}0.892}} & \textbf{{\color{blue}0.863}}\\

				\textbf{UFO \textit{(Ours)}} & VGG16-224 & \ding{52} & U & 0.01 & \textbf{{\color{red} 0.015}} & \textbf{{\color{blue}0.918}}  & \textbf{{\color{blue}0.906}}  & \textbf{{\color{blue}0.031}} & \textbf{{\color{blue}0.891}} & \textbf{{\color{blue}0.888}}  & \textbf{{\color{blue}0.013}} & \textbf{{\color{red}0.959}}  & \textbf{{\color{red}0.951}} & \textbf{{\color{red}0.013}} & \textbf{{\color{red}0.899}} & \textbf{{\color{red}0.869}}\\
				\bottomrule %添加表格底部粗线
		\end{tabular}}    
\end{center}\caption{{Comparisons of our method with the other state-of-the-arts on VSOD datasets. The majority of the results are borrowed from~\cite{ji2021full}. \textbf{\textit{OF}} denotes the optical flow. \textbf{\textit{RT}} denotes the runtime (s). “\textbf{U}”: unsupervised method. “\textbf{S}”: semi-supervised method. “Res” denotes ResNet~\cite{he2016deep}, “Mob” denotes Moiblenet~\cite{howard2019searching}, and the number behind them is the input resolution. 
$\dag$ denotes video segmentation methods trained on DAVIS17~\cite{pont20172017} and YouTube-VOS~\cite{xu2018youtube} datasets, whose results are acquired from their released code weights.
The best two results on each dataset are shown in {\color{red} red}, {\color{blue} blue}.}}\label{table8}
\end{table*}

\noindent \textbf{Co-Saliency Detection.} Likewise, Tab~\ref{table7} presents the comparisons of our method with other state-of-the-arts in CoSD. Note that all methods use the VGG16 as backbone and the input size is 224 $\times$ 224 for fair comparisons unless otherwise specified. We can observe that our framework can also outperform other methods except for two second-best (\ie $S_m$ = 0.860 in Cosal2015 and $MAE$ = 0.073 in CoSOD3k) results, and all the rest results are ranked the first.
It is worth mentioning that unlike some of the previous methods (\ie GCAGC~\cite{zhang2020adaptive} and CADC~\cite{zhang2021summarize}), we do not require additional data~\cite{wang2017learning,liu2010learning} and data augmentation strategies like Jigsaw in GICD~\cite{zhang2020gradient} and copy-and-paste in CACD~\cite{zhang2021summarize} to train the network. Moreover, we visualize the results of some examples (\ie $\textit{pumpkin}$ and $\textit{soap}$) with complex backgrounds and some foreground distractors in Fig~\ref{fig6} top. Our method can yield more accurate co-object masks compared to other approaches, which validates the robustness and effectiveness of our method. Fig~\ref{fig6} bottom shows the PR and the ROC curves of the compared methods, and our curves are higher than the other methods on the three challenging benchmarks.

\vspace{1ex}

\noindent \textbf{Video Salient Object Detection.} We further evaluate our method on VSOD benchmarks. As is shown in Tab~\ref{table8}, our proposed framework can once again achieve the best results on FBMS, ViSal and SegV2 datasets by using the VGG16 backbone and small input resolution but without optical flow.
This shows that our method can effectively extract the object's spatial appearance information and long-range temporal dependencies, which can also reach 140 FPS for real-time inference.
Moreover, the proposed framework is an \textbf{unsupervised} VSOD method, which can outperform some \textbf{semi-supervised} methods~\cite{park2021learning,mei2021transvos,yan2019semi} without any bells and whistles.
The reason why we fail to achieve the best results on the DAVIS is that in some cases the background of co-occurring objects is incorrectly segmented (\eg foreground: the dancing girl; background: the audience. They all belong to the object of “people”). 
To alleviate this issue, we combine the flow cues to make our network pay more attention to the foreground moving salient objects and using optical flow in our network is straightforward.
Specifically, the images and their corresponding optical flow are passed through the share-encoder at the same time, and the enhanced image features to be fed into subsequent networks can be obtained by multiplying the flow features.
More details of the usage of optical flow can be referred to the supplementary material.
%we adopt the copy-and-paste data augmentation strategy from~\cite{zhang2021summarize}, which enables the network to pay more attention to the foreground salient objects. 
%We then can achieve new satisfactory performances on DAVIS ($MAE$ = 0.016 ranks 1$^{st}$; $S_m$ = 0.909 ranks 2$^{nd}$; $F_{\beta}$ = 0.887 ranks 3$^{rd}$).
The quantitative results show that we can achieve new satisfactory results. The performances on FBMS drop a little since the acquired optical flow information are poor.
Note that using the optical flow is not our framework's main purpose and contribution, we recommend using the multi-frame images alone, which can already achieve state-of-the-art performance in most tasks.
Some qualitative results on the 4 VSOD datasets are shown in Fig~\ref{fig7}.
%Note that our method is not specifically designed for VSOD, there may be some limitations in motion attention module that are not as good as some previous methods. Our proposed unified framework can jointly perform CoS, CoSD and VSOD tasks and achieve new state-of-the-art level, which is acceptable. 

\section{Conclusion}
In this paper, we propose a unified framework for group-based image segmentation, which can conduct co-segmentation, co-saliency detection and video salient object detection tasks. The proposed transformer block and intra-MLP module can both help the network well capture the inter-collaborative and intra-activated information. The competitive results on 11 benchmarks validate the effectiveness of our method.
We take an early attempt to unify the deep learning network for different tasks, and we hope this finding will encourage the development of more follow-up research that simplifies the network in other domains.

\section*{Acknowledgment}

This work was supported by National Natural Science Foundation of China (NSFC) 61876208, Key-Area Research and Development Program of Guangdong Province 2018B010108002, and the National Research Foundation, Singapore under its AI Singapore Programme (AISG Award No: AISG-RP-2018-003), and the MOE Tier-1 research grants: RG28/18 (S), RG22/19 (S) and RG95/20.

%\section*{Acknowledgment}


\ifCLASSOPTIONcaptionsoff
  \newpage
\fi

\bibliographystyle{IEEEtran}
\bibliography{mybib}

\clearpage
\renewcommand\appendixname{}
\appendix
\vspace{-15pt}
\noindent \Large Supplementary Material
\normalsize

\vspace{1ex}

This supplementary material contains more details of the network, including the details of using optical flow, more training details, more analysis, more downstream applications and more additional qualitative results in our paper “A Unified Transformer Framework for Group-based Segmentation: Co-Segmentation, Co-Saliency Detection and Video Salient Object Detection".

\subsection{Details of Using Optical Flow}
\begin{figure}[t]
	\begin{center}
		\centering
		\includegraphics[width=0.5\textwidth]{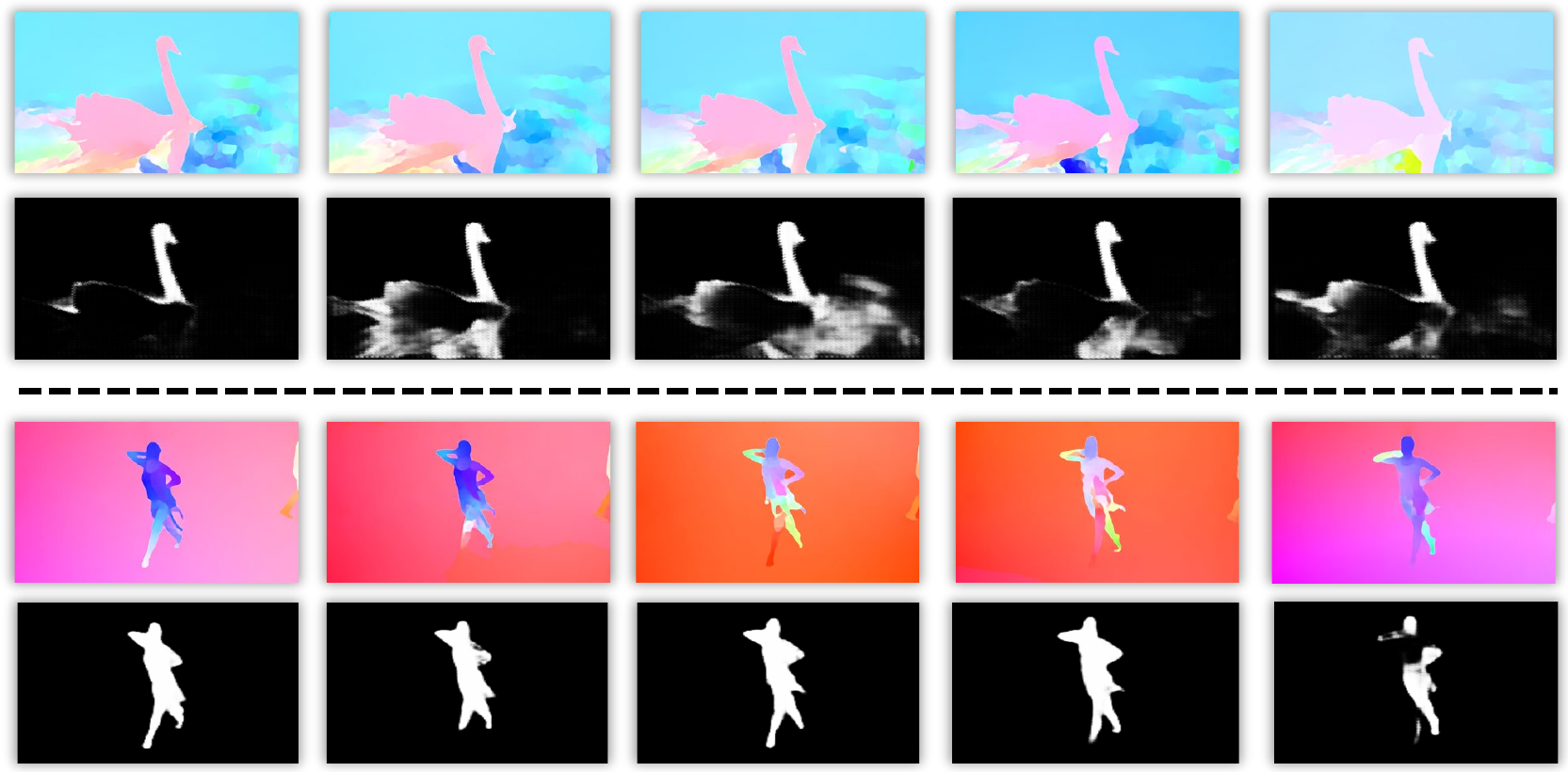}
	\end{center}
	\caption{\textbf{The predictions of the direct optical flow input}.}
	\label{figA}
	%\vspace{-3ex}
\end{figure}

\begin{figure*}[t]
		\begin{center}
			\centering
			\includegraphics[width=6.8in]{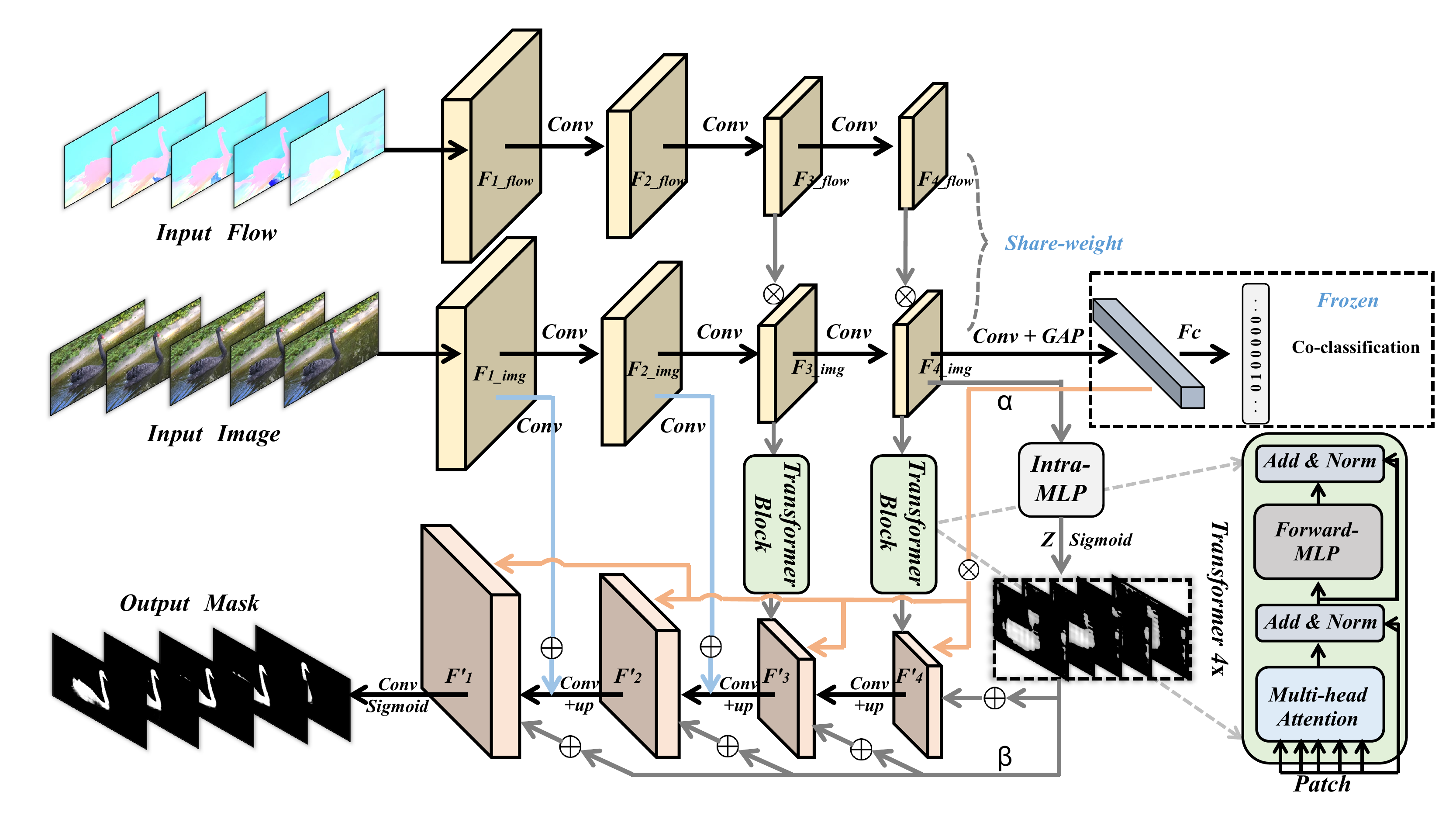}
		\end{center}
		\caption{\textbf{The overall architecture of combing the optical flow information}.}
		\label{figB}
\end{figure*}

As we have discussed in our main paper, optical flow is optional to add into our network, and applying flow cues to our network is straightforward.
First of all, let us first introduce why optical flow can help us. As is
shown in Fig~\ref{figA}, when we directly use the optical flow as input, the output predictions are comparable. This is because the optical flows reflect the foreground moving salient objects without any complicated backgrounds, and thus, our network can well extract their features and yields the masks. 
Based on this, we consider that combing the optical flow features with the origin image feature can help the network focus on more foreground salient objects. Fig~\ref{figB} shows us the overall architecture of using the optical flow information.
Specifically, we maintain the transformer block, intra-MLP learning module, and decoder unchanged but slightly modify the encoder features. We take both the optical flows generated by FlowNet2.0~\cite{ilg2017flownet} and adjacent frames as input, and they share the same encoder.
The enhanced features of the group-based images to be fed into the subsequent network can be formulated as:

\begin{equation}
%\footnotesize
\begin{split}
F_{3\_img} &= F_{3\_flow}\times F_{3\_img}, \\
F_{4\_img} &= F_{4\_flow}\times F_{4\_img}.
\end{split}
\tag{8}
\label{eq8}
\end{equation}

This operation enables the image features to pay more attention to the targeted regions by combining flow cues regardless of the distracting background items.
Note that our method of using optical flow may not be the optimal way, maybe using a separate optical flow branch and a unique combination can further improve the network performance. However, using the optical flow is not our framework's main purpose and contribution, we can already achieve state-of-the-art performance on three benchmarks (\ie FBMS~\cite{ochs2013segmentation}, ViSal~\cite{wang2015consistent} and SegV2~\cite{li2013video}) without using it.
We will continue to refine and research possible ways to improve our network in future work.

\subsection{More Training Details}
For object co-segmentation (CoS) and co-saliency detection (CoSD), unlike the previous methods (\ie GCAGC~\cite{zhang2020adaptive}, CADC~\cite{zhang2021summarize}) that are trained on both COCO-SEG~\cite{lin2014microsoft} and other saliency datasets (\ie MSRA-B~\cite{liu2010learning}, DUTS~\cite{wang2017learning}),
we train our method only using COCO-SEG~\cite{lin2014microsoft} dataset and without using data augmentation like copy-and-paste in CADC~\cite{zhang2021summarize}, Jigsaw in GICD~\cite{zhang2020gradient}.
For video salient object detection (VSOD), we follow the previous works (\ie FSNet~\cite{ji2021full}, PCSA~\cite{gu2020pyramid}) that employ the random flip, random rotation and multi-scale $\left\{ 0.75, 1, 1.25 \right\}$ training strategy in the training phase.

\subsection{More Analysis}

\noindent \textbf{The number of $K$.} We here explore the effect of the $K$ number we use in the intra-MLP learning module. As shown in Tab~\ref{tableA}, we find that when $K = 4$, our network can achieve the best performance. We conjecture that a small or too large $K$ number may make the network learn less useful corresponding semantic information or redundant information that harms the network.

\vspace{1ex}

\begin{table}[t]
    \centering
   
            \scalebox{1.15}{
                \begin{tabular}{c|cc|cc}
        		\toprule  %添加表格头部粗线
        		\toprule  %添加表格头部粗线
        		\multirow{2}*{\text{Method}} 
        	    &
        		\multicolumn{2}{c|}{PASCAL}  &
			    \multicolumn{2}{c}{iCoseg} \\
        		\cline{2-5}
        		 & $\mathcal{P}$ & $\mathcal{J}$& $\mathcal{P}$ & $\mathcal{J}$ \\
        		
        		\toprule  %添加表格头部粗线
        		\text{$K = 2$} & 94.5 & 72.8 & 97.1 & 90.2 \\
        		\rowcolor{mygray} \textbf{$K = 4$} & \textbf{95.4} & \textbf{73.6} & \textbf{97.6} & \textbf{90.9}\\
        		\text{$K = 6$} & 95.2 & 73.3 & 97.5 & 90.6 \\
        		\bottomrule %添加表格底部粗线
                \end{tabular}
        }
            \vspace{6pt}
            \caption{Analysis of the $K$ number in intra-MLP learning module.}\label{tableA}
		    
\end{table}

\begin{table}[!bt]
    \centering
        \scalebox{1.15}{\begin{tabular}{c|cc|cc}
        		\toprule  %添加表格头部粗线
        		\toprule  %添加表格头部粗线
        		\multirow{2}*{\text{Method}} 
        	    &
        		\multicolumn{2}{c|}{PASCAL}  &
			    \multicolumn{2}{c}{iCoseg} \\
        		\cline{2-5}
        		 & $\mathcal{P}$ & $\mathcal{J}$& $\mathcal{P}$ & $\mathcal{J}$ \\
        		
        		\toprule  %添加表格头部粗线
        		\text{ D = 512} & 94.7 & 72.3 & 96.5 & 89.5 \\
        		\rowcolor{mygray} \textbf{D = 782} & \textbf{95.4} & \textbf{73.6} & \textbf{97.6} & \textbf{90.9}\\
        		\text{D = 1024} & 94.9 & 73.1 & 97.0 & 90.3 \\
        		\bottomrule %添加表格底部粗线
                \end{tabular}
        }
        \caption{Analysis of the dim channel in transformer  projection function.}\label{tableB}
\end{table}

\noindent \textbf{The Dim Channels.} We further conduct an additional experiment to analyze the effect of the dim channel in our transformer block projection function. Since the top-level output features map channels are 512, and thus, we explore $\left\{ D = 512, 782, 1024 \right\}$, respectively. Tab~\ref{tableB} shows that when $D = 782$ the network performs the best. Note that keep enlarging the dim channel will not only bring no gain to the network, but also consume computing resources.

\vspace{1ex}

\begin{table*}[t]
    \renewcommand\arraystretch{1.5}
	\begin{center}
		\scalebox{0.85}{
			\begin{tabular}{c|c|c|c|c|ccc|ccc|ccc|ccc}
				\toprule  %添加表格头部粗线
				\multirow{2}*{Method} & \multirow{2}*{Setting} & \multirow{2}*{\textbf{\textit{OF}}} &
				\multirow{2}*{\textbf{\textit{Sup}}} &
				\multirow{2}*{\textbf{\textit{RT}}} &
				\multicolumn{3}{c|}{DAVIS\text{$_{16}$}} & \multicolumn{3}{c|}{FBMS} &  \multicolumn{3}{c|}{ViSal} &
				\multicolumn{3}{c}{SegV2}
				\\
				\cline{6-17}
				{~}&{~}&{~}&{~}&{~}&
				\textit{MAE$\downarrow$} & \textit{$S_m\uparrow$} & \textit{$F_{\beta}\uparrow$} &
				\textit{MAE$\downarrow$} & \textit{$S_m\uparrow$} & \textit{$F_{\beta}\uparrow$} &
				\textit{MAE$\downarrow$} & \textit{$S_m\uparrow$} & \textit{$F_{\beta}\uparrow$} &
				\textit{MAE$\downarrow$} & \textit{$S_m\uparrow$} & \textit{$F_{\beta}\uparrow$}  \\
				% \cmidrule{2-7} & ZSL \ &  GZSL \ &  ZSL \ & GZSL \ &  ZSL \ &  GZSL \ \\
				\toprule  %添加表格头部粗线
				
				%  \cellcolor{white}
				
				\textbf{UFO \textit{(Ours)}} & HRNet-224 & \XSolidBrush& U & 0.01 & 0.028 & 0.881  & 0.842 & 0.026 & 0.899 & 0.893  & 0.012 & 0.958  & 0.948 & 0.018 & 0.897 & 0.866\\
	
				\textbf{UFO \textit{(Ours)}} & HRNet-224 & \ding{52} & U & 0.03 & 0.013 & 0.921  & 0.907  & 0.033 & 0.888 & 0.887  & 0.011 & 0.962  & 0.956 & 0.012 & 0.901 & 0.867\\
				\bottomrule %添加表格底部粗线
		\end{tabular}}    
\end{center}\caption{\textbf{\textit{OF}} denotes the optical flow. \textbf{\textit{RT}} denotes the runtime (s). “\textbf{U}”: unsupervised method. The number behind the backbone is the input resolution.}\label{tableC}
%\vspace{-6ex}
\end{table*}

\noindent \textbf{Stronger Backbone.} Furthermore, we adopt a stronger backbone HRNet~\cite{sun2019deep} to conduct experiments on VSOD tasks, whose results are shown in Tab~\ref{tableC}. Although using a more powerful backbone can improve the performance of the network, the execution time will also increase at the same time. 
As we mentioned in our main paper, using vgg16 backbone can already achieve the state-of-the-art performance and the execution speed can also reach real-time. Therefore, we recommend using vgg16, which can already meet most of the needs in our tasks.

\subsection{More Downstream Applications}

\noindent \textbf{Co-Localization.} We try to conduct an additional experiment on CUB-2011~\cite{wah2011caltech} dataset for co-localization task, which is also aims to simultaneously localize  objects of the same class across a set of distinct images~\cite{tang2014co}. 
%We compare our method with some state-of-the-art approaches in co-segmentation task (\ie SSNM~\cite{zhang2020deep}) and co-saliency detection task (\ie CADC~\cite{zhang2021summarize}, GCAGC~\cite{zhang2020adaptive}).
%For a fair comparison, xxx
As shown in Fig~\ref{figD}, we first generate the object masks on the co-objects. Then, we obtain the green predated box according to the binary masks. As can be seen, our predicted bounding boxes are closed to the GT bounding boxes.

\vspace{1ex}

\noindent \textbf{Real-Time Video Salient Object Detection.}
Our framework can be applied to detect the salient object in the video without using optical flow, therefore, it can be easily to conduct this task. We use a video sample from the public website\footnote[1]{ \url{https://www.youtube.com/watch?v=aSqeWUuQSlM&t=1s}} for our demo. The result are provided in our publicly released code page. Note that we do not train our network on YouTube-VOS~\cite{xu2018youtube} dataset, and the competitive result of the randomly selected demo video shows the 
superiority of our proposed method. 

\vspace{1ex}

\noindent \textbf{Bullet-Chat Blocking.} Bullet-chat blocking aims to prevent the salient objects in the video from being occluded by the viewer's barrage. It has to segment out the object and yield a accurate mask that provides to block the bullet-chat text in the foreground.
To validate the generality of our method, we use a video sample from the public website\footnote[2]{\url{https://www.youtube.com/watch?v=kpUYkG0FAYk&t=146s}}. And then we adopt the OpenCV~\cite{bradski2000opencv} toolbox to generate some pseudo bullet-chat that constantly slides over the video. 
Finally, we can produce a new bullet-chat blocking video as we provide in our publicly released code page.

\begin{figure}[t]
		\begin{center}
			\centering
			\includegraphics[width=3.3in]{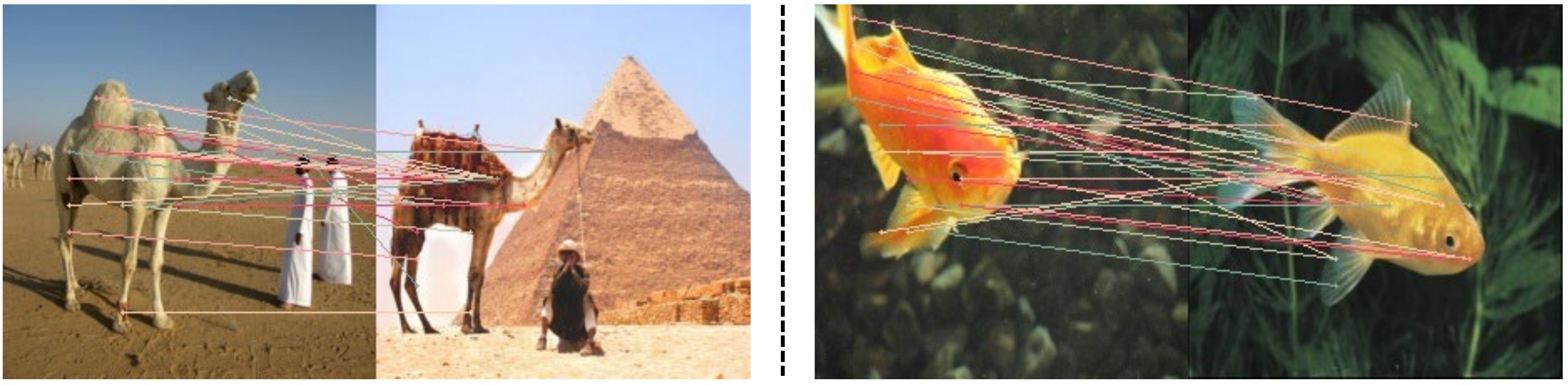}
		\end{center}
		\caption{\textbf{Visualizations of object matching flow}.}
		\label{figC}
\end{figure}

\begin{figure*}[t]
		\begin{center}
			\centering
			\includegraphics[width=6.8in]{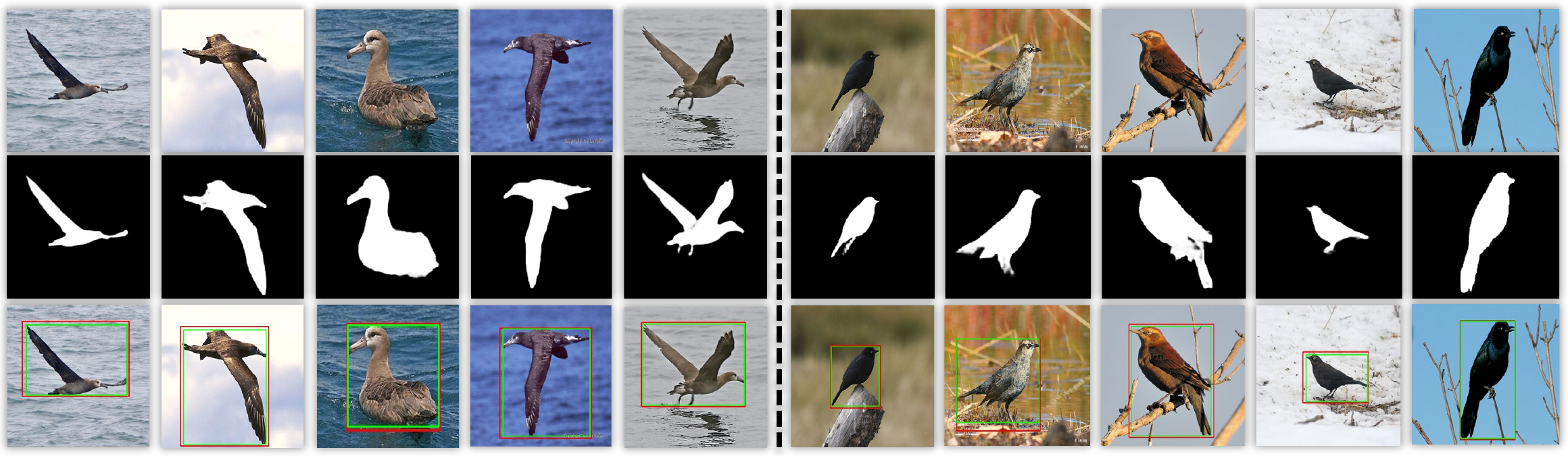}
		\end{center}
		\caption{\textbf{Visualizations of co-localization.} The 1$^{st}$ row are the input images, the 2$^{nd}$ row are our predicted masks, and the 3$^{rd}$ row are the localization results. The ground-truth bounding box is in {\color{red} red} and predicted bounding box is {\color{green} green}.}
		\label{figD}
\end{figure*}

\subsection{More Visualizations}

\noindent \textbf{Matching.}
To further provide the network interpretability, we visualize how group-based images match features with each other during network learning.
Fig~\ref{figC} shows the highly correlated patches in different images, which illustrates that our network can well capture the object similarities and discover the object regions. 

\vspace{1ex}

\begin{figure*}[t]
		\begin{center}
			\centering
			\includegraphics[width=5.5in]{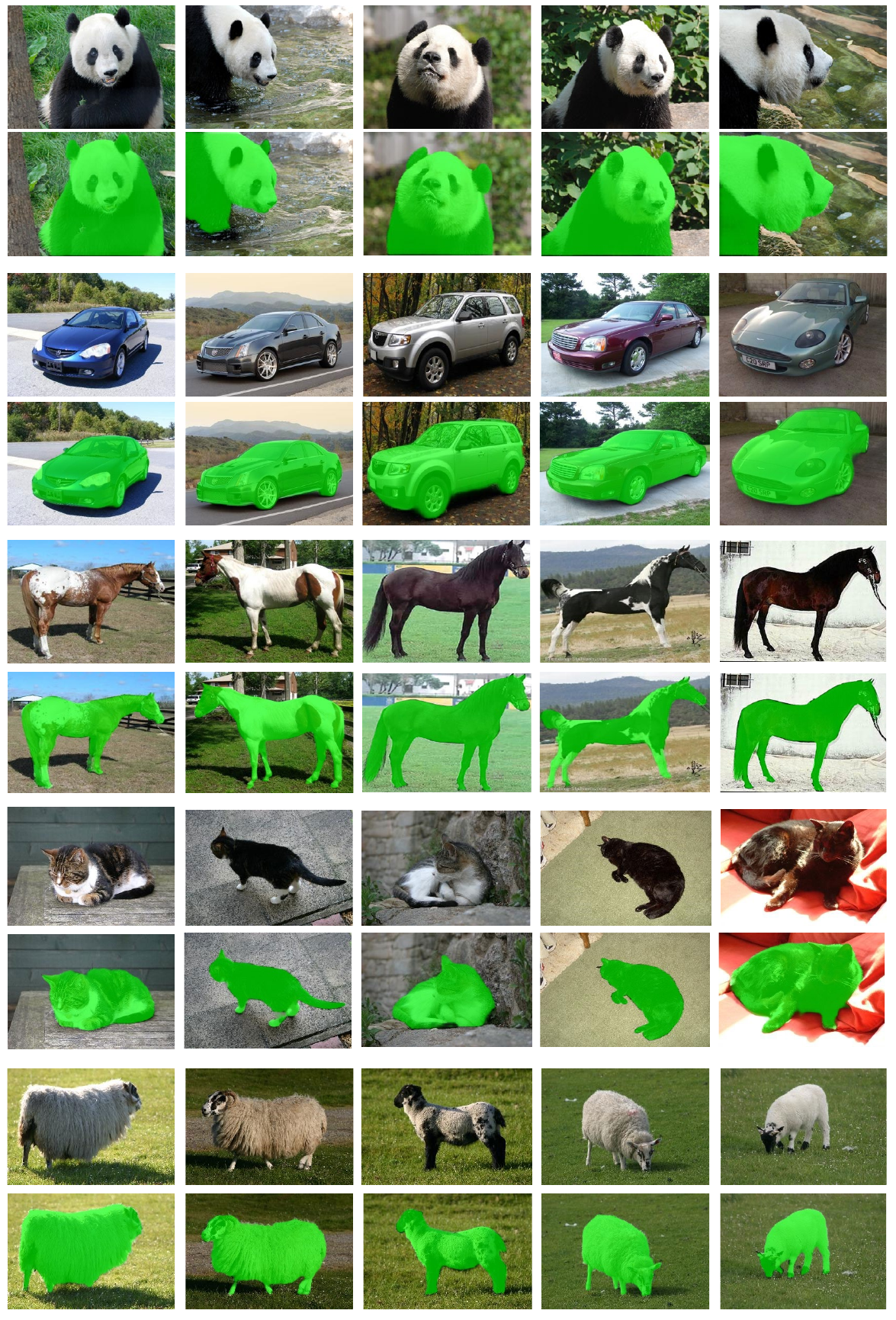}
		\end{center}
		\caption{\textbf{Visualizations of co-segmentation}.}
		\label{figs}
\end{figure*}

\noindent \textbf{Co-Segmentation.}
Fig~\ref{figs} shows more visualizations on CoS task.

\vspace{1ex}

\begin{figure*}[t]
		\begin{center}
			\centering
			\includegraphics[width=5.5in]{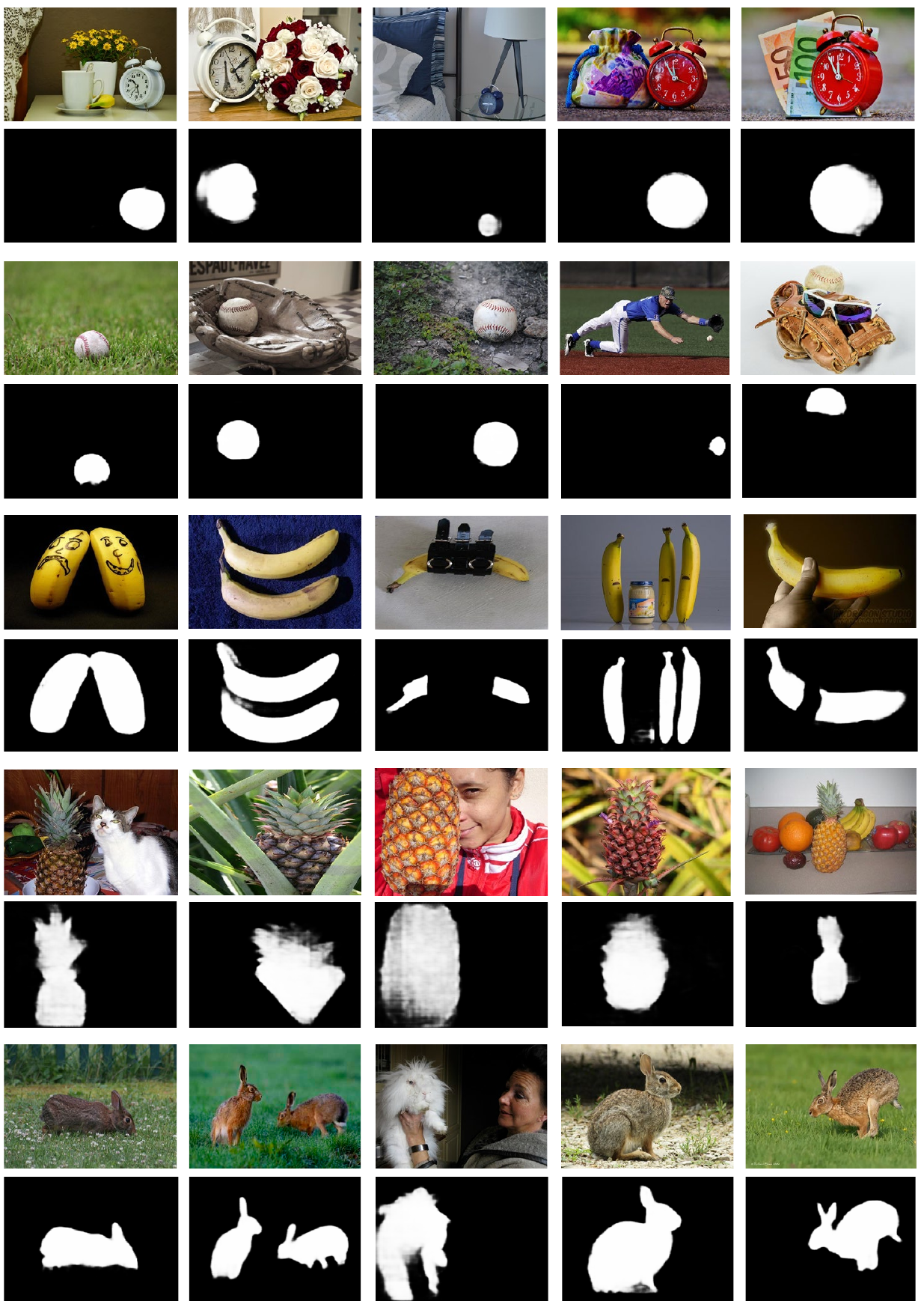}
		\end{center}
		\caption{\textbf{Visualizations of co-saliency detection}.}
		\label{figy}
\end{figure*}

\noindent \textbf{Co-Saliency Detection.}
Fig~\ref{figy} shows more visualizations on CoSD tasks.

\vspace{1ex}

\begin{figure*}[t]
		\begin{center}
			\centering
			\includegraphics[width=2.8in]{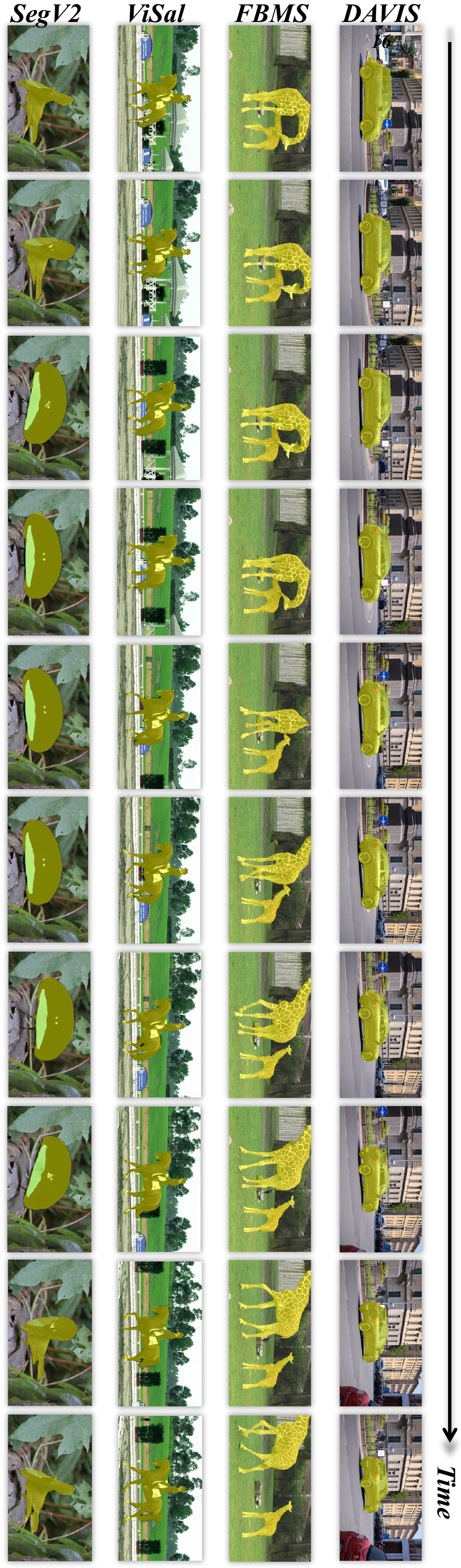}
		\end{center}
		\caption{\textbf{Visualizations of video salient object detection}.}
		\label{figk}
\end{figure*}

\noindent \textbf{Video Salient Object Detection.}
Fig~\ref{figk} shows more visualizations on VSOD tasks.

\vspace{1ex}

\noindent Note that all the predicted results can be found and downloaded in our project page: \url{https://github.com/suyukun666/UFO}.

\end{document}